\documentclass{article}

\PassOptionsToPackage{numbers, compress}{natbib}
\usepackage[preprint]{neurips_2026}

\usepackage[utf8]{inputenc} 
\usepackage[T1]{fontenc}    
\usepackage{hyperref}       
\usepackage{url}            
\usepackage{booktabs}       
\usepackage{amsfonts}       
\usepackage{nicefrac}       
\usepackage{microtype}      
\usepackage{xcolor}         

\usepackage{amsmath}
\usepackage{array}
\usepackage{booktabs}
\usepackage{makecell}
\usepackage{graphicx}
\usepackage{caption}
\usepackage{placeins}
\usepackage{multirow}
\usepackage{subcaption}

\newcommand{\etal}{\textit{et al.}}

\title{Sketch2MinSurf: Vision-Language Guided Generation of Editable Minimal Surfaces from Hand-Drawn Sketches}

\author{%
  Wenda Wang$^{1,}$\thanks{Equal contribution.} \quad 
  Anqi Liu$^{2,}$\footnotemark[1] \quad 
  Junqi Yang$^{1}$ \quad 
  Lei He$^{1}$ \\
  \textbf{Luying Wang$^{1}$ \quad Jiachen Lu$^{1}$ \quad Weixin Huang$^{1,}$\thanks{Corresponding author.}} \\
  \normalfont $^{1}$School of Architecture, Tsinghua University \\
  \normalfont \texttt{\{wwd23, yang-jq23, helei23, wangly21, lu-jc21\}@mails.tsinghua.edu.cn} \\
  \normalfont \texttt{huangwx@tsinghua.edu.cn} \\
  \normalfont $^{2}$Department of Architecture, National University of Singapore \\
  \normalfont \texttt{anqiliu@u.nus.edu}
}

\begin{document}

\maketitle

\begin{abstract}
Converting hand-drawn sketches into structured 3D geometries remains challenging due to the difficulty of representing non-Euclidean surfaces and maintaining topological consistency. Existing generative models such as GANs, NeRFs, and diffusion architectures often fail to produce editable manifolds directly usable in downstream design workflows. We present \textbf{Sketch2MinSurf}, a hybrid vision-language and geometric optimization framework that integrates vision–language guidance with minimal-surface theory to generate smooth and editable 3D surfaces from hand-drawn sketches. The core of our approach is a spatial–topological encoding that represents geometry as tuples of node coordinates and real/virtual edge skeletons, enabling stable topological control during generation. We further introduce the Sketch2MinSurf Structural Loss (S2MS-Loss), a reward-modulated objective that jointly constrains geometric reconstruction and topological coherence. On a test set of 100 sketches, \textbf{Sketch2MinSurf} achieves a topological similarity score of 0.844, outperforming existing sketch-to-shape baselines. The generated manifolds are directly editable and free from non-manifold artifacts. A public art installation at a university showcases the method’s potential for human-intent-driven 3D form generation. The dataset and code are available at \url{https://anonymous.4open.science/r/Sketch2MinSurf/}.
\end{abstract}

\section{Introduction}
\label{sec:Introduction}

With the growing demand for efficiency, creativity, and intelligence in architectural design, generative AI--based assistive technologies have emerged as a promising research frontier. Architects, clients, and the public increasingly favor freeform and minimal-surface geometries in large-scale works and public installations. However, modeling such non-Euclidean forms remains a fundamental challenge: architects must handle complex spatial-topological relationships, ambiguously defined parameters, and difficulties in reconstructing intended geometries with high fidelity.

Hand-drawn sketches---a fast, intuitive, and flexible design medium---remain indispensable during early conceptualization. Yet, the implicit spatial cues and geometric constraints embedded in 2D sketches are difficult to infer in 3D. Automated reconstruction from sketches often introduces spatial distortion and degraded continuity, requiring extensive manual refinement. These limitations not only increase design workload but also obstruct faithful realization of design intent in downstream modeling and fabrication.

\begin{figure}[htbp]
  \centering
  \includegraphics[width=0.6\linewidth]{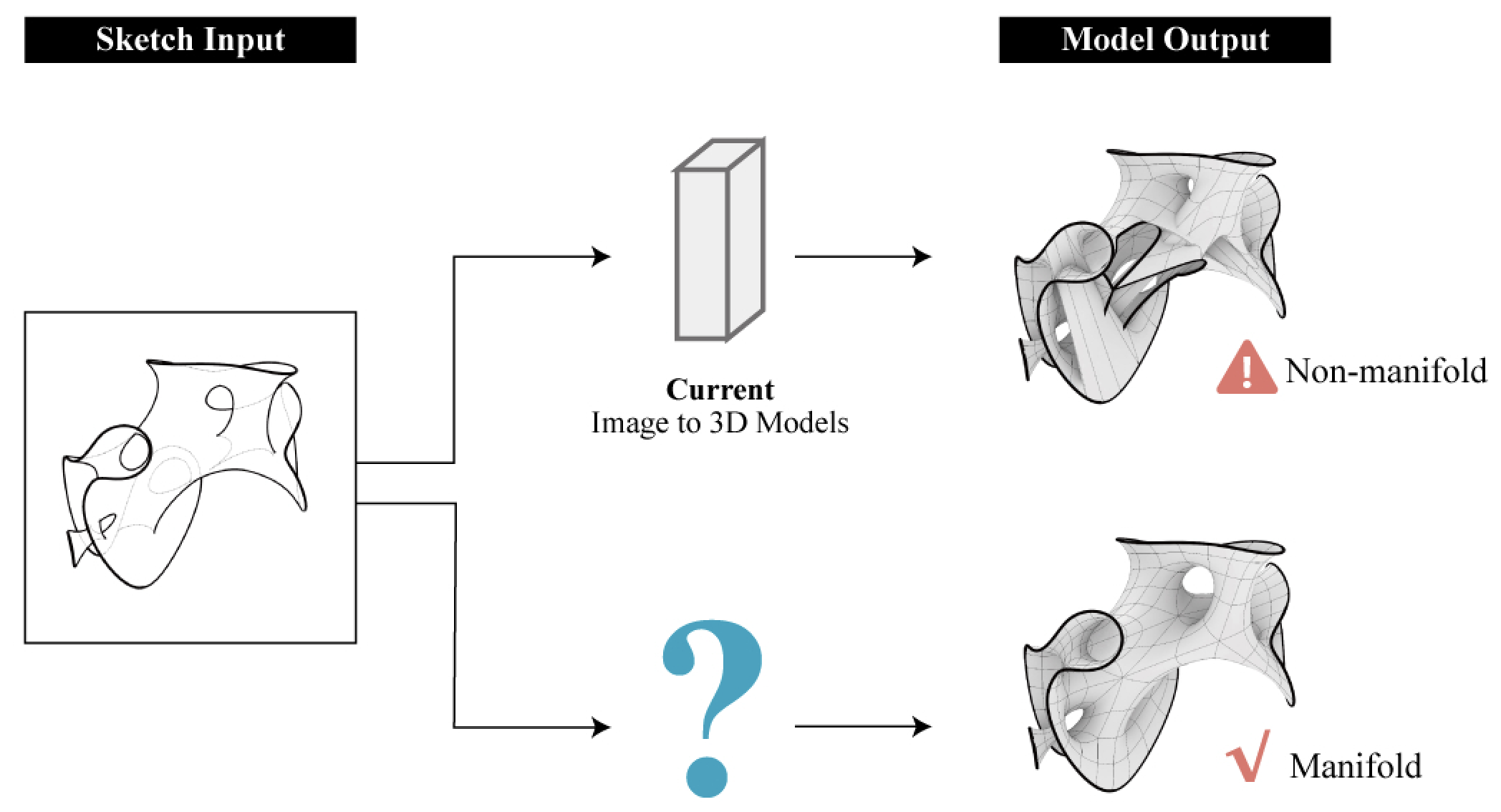}
  \caption{Existing image-to-3D pipelines often yield non-manifold or disconnected meshes, limiting editability and fabrication readiness.}
  \label{fig:current_methods}
\end{figure}

Traditional architectural workflows proceed sequentially from sketches to coarse 3D massing and refined models. This transition from 2D to 3D entails cross-modality reasoning that demands expert knowledge and significant time. Consequently, reconstructing accurate 3D geometry from sketches has become a central challenge at the intersection of architecture and computer vision. Recent frameworks such as CityCraft~\cite{kim2020citycraft}, Vitruvio~\cite{tono2024vitruvio}, and BuilDiff~\cite{wei2023buildiff} leverage GANs~\cite{goodfellow2020gan} and SAM~\cite{kirillov2023segmentanything} to automate sketch-based mass generation. However, most methods lack explicit topological reasoning and produce geometries with non-manifold artifacts (Fig.~\ref{fig:current_methods}).

Recent advances such as NeRF~\cite{mildenhall2021nerf} and Gaussian Splatting~\cite{kerbl2023gaussiansplatting} have significantly improved view synthesis and volumetric scene representations. However, these methods remain constrained by implicit radiance structures that are difficult to edit or convert into clean and usable meshes. Diffusion-based techniques, including DiffRF~\cite{muller2023diffrf}, Zero123++~\cite{shi2023zero123pp}, Direct3D~\cite{wu2024direct3d}, and Unique3D~\cite{wu2024unique3d}, can produce visually compelling renderings, yet they often fail to ensure watertight and manifold topology, which is essential for applications in architecture and fabrication.

In parallel, emerging multimodal vision–language models (VLMs) such as SpatialVLM~\cite{chen2024spatialvlm}, SpatialPIN~\cite{ma2024spatialpin}, and MetaSpatial~\cite{pan2025metaspatial} have begun to integrate spatial reasoning capabilities. Despite this progress, their performance still relies heavily on dataset-specific priors, and they exhibit limited understanding of abstract geometric relationships. This shortcoming becomes particularly critical when processing freehand architectural sketches, where spatial constraints must be inferred rather than retrieved from data. The difficulty is further amplified in minimal-surface generation tasks that demand both topological consistency and geometries suitable for physical fabrication.

To address these challenges, we propose \textbf{Sketch2MinSurf}, a hybrid vision-language and geometric optimization framework that bridges \textbf{design intent, spatial reasoning, and geometric generation}. Our method decouples the pipeline into a topology-aware VLM prediction stage and a geometric solver stage, transforming hand-drawn sketches into editable, manifold 3D surfaces suitable for architectural design and fabrication. Experiments across five VLM architectures show that the proposed reward-modulated training objective consistently improves topological fidelity, achieving a topological similarity score of 0.844. Specifically, our contributions are threefold:

\begin{itemize}
    \item Topology-aware skeleton encoding. We introduce a structured representation that interprets sketch-embedded design intent as parametric 3D skeletons, maintaining topological coherence throughout generation.
    \item Reward-modulated structural loss. We propose S2MS-Loss, which evaluates skeleton quality across five structural dimensions and uses the aggregated accuracy to dynamically scale the cross-entropy loss, guiding the model toward topologically valid outputs. By decoupling VLM prediction from geometric solving, the pipeline maintains full editability and enforces strict fabrication constraints that purely neural representations cannot guarantee.
    \item Editable and manifold output. The generated minimal surfaces are inherently watertight and parametrically controllable, enabling design iteration and direct fabrication integration.
\end{itemize}

By integrating architectural reasoning with state-of-the-art neural 3D generation, \textbf{Sketch2MinSurf} advances cross-disciplinary research at the intersection of generative AI, computational geometry, and architectural design.

\section{Related Work}
\label{sec:RelatedWorks}

Generative 3D modeling from limited or abstract inputs has emerged as a central theme across computer vision, graphics, and architectural design. Our research connects three complementary directions, namely \textit{image-to-3D generation}, \textit{vision–language spatial reasoning}, and \textit{minimal-surface geometry}, to form a unified and design-aware reconstruction pipeline.

\subsection{Image-to-3D generation}
Recent years have seen rapid progress in neural 3D generation, spanning NeRF-based~\cite{mildenhall2021nerf} and diffusion-based~\cite{ho2020ddpm} approaches~\cite{melas2023realfusion,nichol2022pointe,cheng2023sdfusion,szymanowicz2023viewsetdiffusion}. More recent efforts such as Direct3D~\cite{wu2024direct3d} and Unique3D~\cite{wu2024unique3d} leverage native 3D latent diffusion to improve generation fidelity and efficiency from single images. Although these methods produce detailed geometry, they often yield implicit or non-manifold structures that are difficult to edit for downstream design tasks.

Within architectural design, sketch-based 3D generation methods such as CityCraft~\cite{kim2020citycraft}, Vitruvio~\cite{tono2024vitruvio}, and BuilDiff~\cite{wei2023buildiff} generate coarse volumetric forms but remain confined to Euclidean typologies, struggling with complex topologies or minimal geometries that require continuous surfaces and precise curvature control.

\subsection{Vision-language models and spatial reasoning}
The contrastive pretraining framework introduced by CLIP~\cite{radford2021clip} established a shared embedding space for images and text, providing the foundation for large vision–language models (VLMs) such as BLIP-2~\cite{li2023blip2}, GPT-4V~\cite{openai2023gpt4}, and Qwen2-VL~\cite{wang2024qwen2vl}. These models have begun to exhibit cross-modal reasoning capabilities, yet they still represent geometry implicitly rather than explicitly.

In recent years, spatial reasoning has become a dedicated research frontier. SpatialVLM~\cite{chen2024spatialvlm} and SpatialRGPT~\cite{cheng2024spatialrgpt} introduced large-scale 3D reasoning benchmarks, SpatialPIN~\cite{ma2024spatialpin} enhanced spatial reasoning through prompting and interacting with 3D foundation model priors, while LayoutVLM~\cite{sun2025layoutvlm} and MetaSpatial~\cite{pan2025metaspatial} focused on layout generation and user-driven spatial constraint learning. SpatialCoT~\cite{liu2025spatialcot} further integrated coordinate alignment with chain-of-thought reasoning to enhance geometric awareness. Despite these advancements, current VLMs do not support explicit geometric synthesis. Although they demonstrate a degree of spatial understanding, they cannot directly output parameterized or controllable 3D surfaces suitable for real-world design workflows.

Our research diverges from prior work by embedding architectural geometry priors into the learning process. Through minimal-surface regularization and parametric topology encoding, our pipeline enables interpretable and editable 3D outputs that closely link symbolic reasoning with geometric form.

\subsection{Minimal surfaces and parametric geometry}
Minimal surfaces, characterized by zero mean curvature, represent locally area-minimizing configurations. Foundational studies, ranging from Douglas’s solution to the Plateau problem~\cite{douglas1939minimalsurfaces} to graph-theoretic formulations by Sadoc and Charvolin~\cite{sadoc1989infinite}, established the mathematical basis for their classification. Later contributions connected computational mathematics with geometric modeling: Andersson \etal~\cite{andersson1988minimalsurfaces} explored Chebyshev normalization methods, and Wang~\cite{wang2009minkowski} formulated potential-field-based isosurface approaches to describe complex curvature behavior.

More recent studies focus on parameterized and generative control of minimal surfaces. Oka \etal~\cite{oka2015transformation} proposed localized transformation methods, Chen \etal~\cite{chen2020liquidframeworks} developed dual-network correspondences for triply periodic structures, and Xiang \etal~\cite{xiang2023blockcopolymer} achieved quantitative curvature modulation through symmetry-based regularization.

In architectural contexts, minimal surfaces serve as a fundamental design typology that combines aesthetic fluidity, structural efficiency, and fabrication feasibility. Built projects such as the \textit{Metropol Parasol}, the \textit{ICD/ITKE Research Pavilion}, and Zaha Hadid Architects’ \textit{Winton Gallery} demonstrate how geometry and structure can be seamlessly unified. Nevertheless, practical realizations frequently rely on ``generalized’’ minimal surfaces~\cite{kapfer2011tissue,fogden1999transformations}, which allow controlled curvature deviations to accommodate material and fabrication constraints. Current computational pipelines rarely offer differentiable control over such manifold geometries, restricting their integration with downstream analysis and fabrication workflows. Our framework addresses this limitation by encoding minimal surface topology as structured skeletal descriptions, enabling VLMs to directly generate parametrically controllable manifold surfaces.
\section{Methodology}

\subsection{Overview of Sketch2MinSurf}

\begin{figure*}[htbp]
  \centering
   \includegraphics[width=0.9\linewidth]{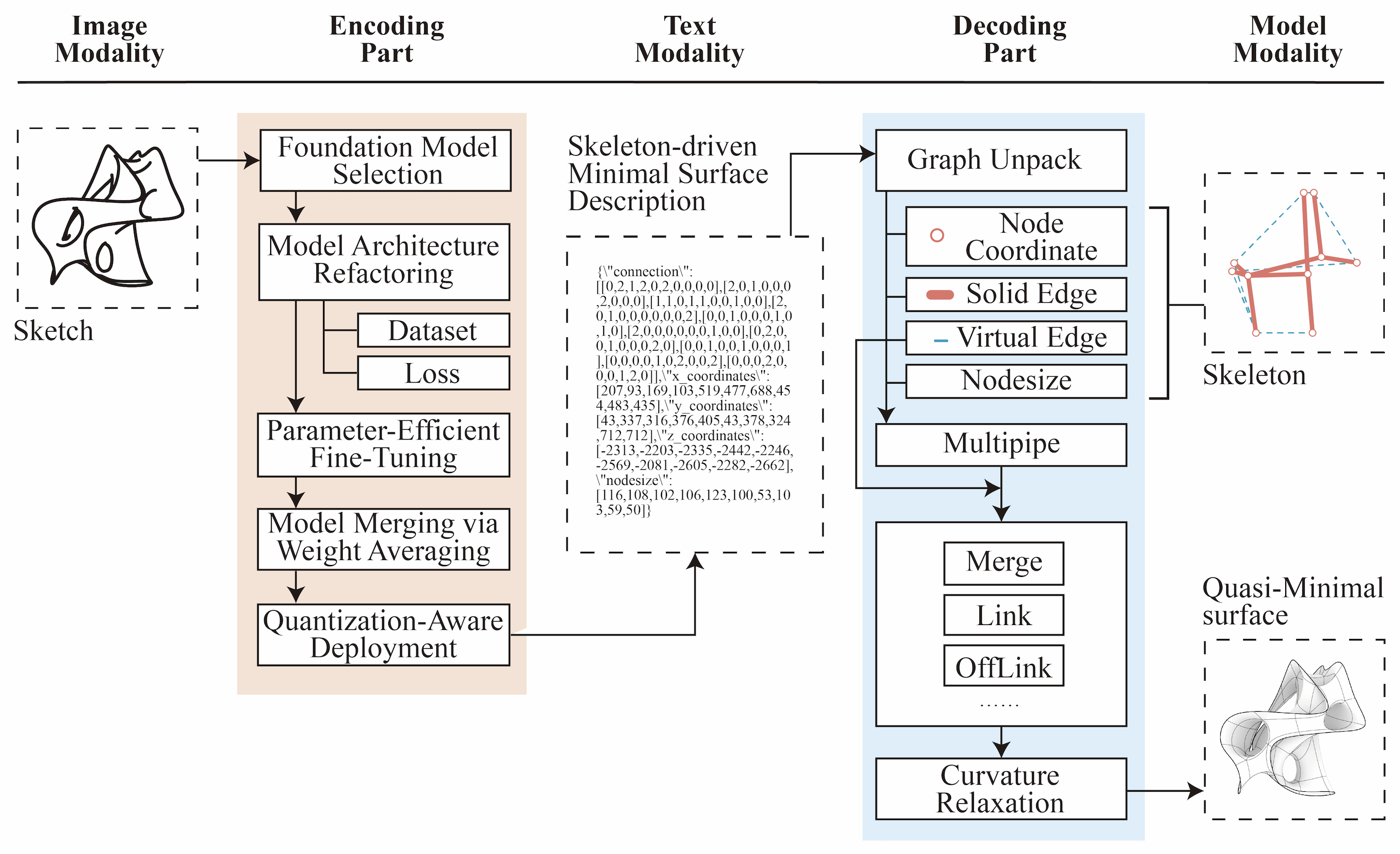}
   \caption{Framework of Sketch2MinSurf.}
   \label{fig:framework}
\end{figure*}

We propose Sketch2MinSurf, a framework that transforms 2D hand-drawn sketches into 3D minimal surfaces through a topology-aware encoding and decoding pipeline (Fig.~\ref{fig:framework}). It consists of three components: (1) a Sketch2MinSurf Encoder that represents minimal surfaces as structured skeleton descriptions with node coordinates, dimensions, and connectivity; (2) a fine-tuned VLM that detects sketches and outputs serialized skeletal descriptions; and (3) a Sketch2MinSurf Decoder that reconstructs 3D geometry through multi-tube operators, topological operations (merge/link/offset), and curvature relaxation to minimize mean curvature under skeletal constraints. The resulting pipeline is controllable, interpretable, and deployable for practical architectural applications.

\subsection{Topology-aware skeleton encoding}

\subsubsection{Minimal surface representation}

We introduce a topology-aware skeleton encoding that decomposes minimal surfaces into two basic elements: saddle regions capturing negative Gaussian curvature, and cylindrical surfaces providing axial extension. This decomposition enables flexible manipulation through geometric operations while maintaining mathematical properties. The key insight is that solid edges constrain the hard boundaries of saddle patches, defining where surfaces must meet, while virtual edges guide tubular transitions between patches without imposing hard boundaries. This dual-edge mechanism reduces complex 3D manifold topology to a compact skeletal description that the solver can reliably reconstruct.

Building on this foundation, we develop the Sketch2MinSurf Description method (Fig.~\ref{fig:framework3}), which represents 3D topological skeletons through three components:

\begin{figure}[htbp]
  \centering
   \includegraphics[width=0.8\linewidth]{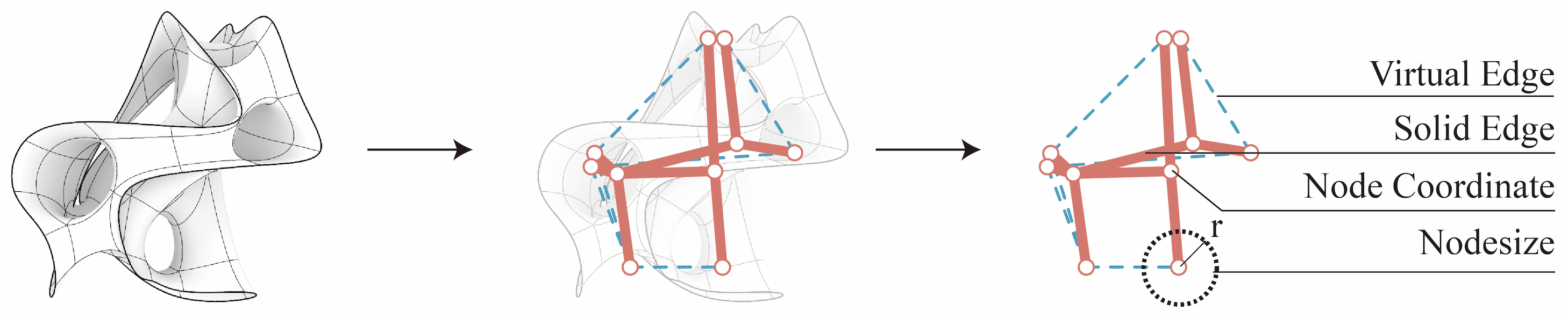}
   \caption{Framework of the Sketch2MinSurf description method.}
   \label{fig:framework3}
\end{figure}

\textbf{Node Information:} Each node $i$ is defined by its 3D coordinates $\mathbf{p}_i = (x_i, y_i, z_i)$ and a size parameter $s_i$ that specifies the local surface offset distance.

\textbf{Connectivity Structure:} We distinguish between two types of edges to capture the complete topological relationships:
\begin{itemize}
\item Solid Edges (SE): Primary connections forming the main topological skeleton along pipe centerlines
\item Virtual Edges (VE): Orthogonal, non-intersecting connections perpendicular to solid edges, constructed using connection operators (Link, Merge, or OffLink)
\end{itemize}

This dual-edge representation enables heterogeneous connections and supports diverse connection logics, significantly enhancing adaptability for complex freeform surface generation (see supplementary Fig.~S3 for an illustration). The skeleton-based approach translates directly into parametric generative workflows while maintaining computational efficiency for large-scale models.

\textbf{Encoding Algorithm:} The Sketch2MinSurf Encoder transforms topological skeletons into 3D minimal surface models by processing node coordinates, SE/VE connectivity, and node size information to generate surfaces with complete internal and external boundaries. The full encoding pipeline is detailed in the supplementary material (Fig.~S4).

\subsubsection{Dataset construction}

We construct a training dataset of 3,000 samples in OpenAI ShareGPT format with image-prompt pairs as input and structured skeletal descriptions as output. Input images are single-view renderings at 750$\times$750 resolution combining line drawings with grayscale shading. The dataset was developed through four iterations (v1.0--v4.0), progressively refining encoding schemes, prompt designs, and visual representations; see the supplementary material for iteration details.

\subsection{Learning objective and training details}

\subsubsection{Sketch2MinSurf structural loss}

\begin{figure}[htbp]
  \centering
   \includegraphics[width=0.9\linewidth]{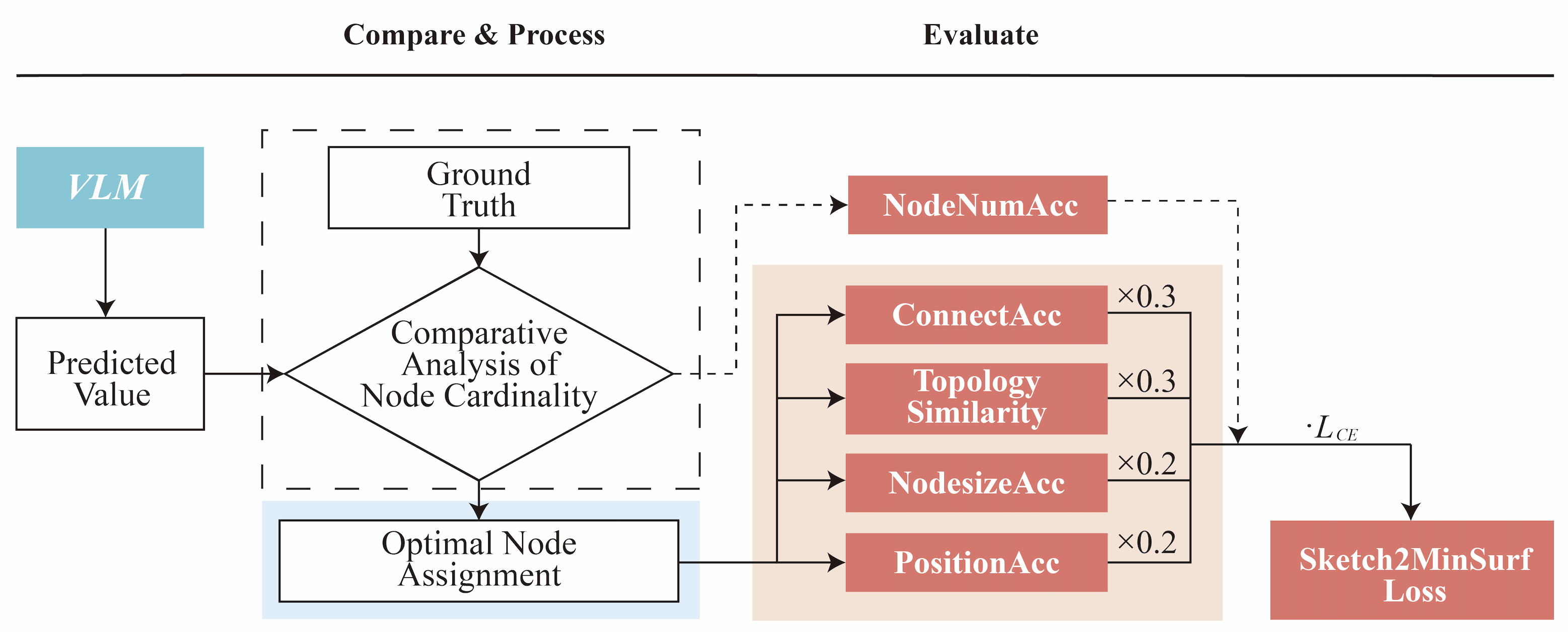}
   \caption{Framework of S2MS-Loss. The structural reward evaluates topological quality and modulates the supervised objective through reward-shaped scaling.}
   \label{fig:framework5}
\end{figure}

We propose the Sketch2MinSurf Structural Loss (S2MS-Loss), which uses structural accuracy as a reward signal to dynamically weight the cross-entropy objective (Fig.~\ref{fig:framework5}).

The S2MS-Loss evaluates skeleton quality across five dimensions:

\textbf{Node Number Accuracy} ($\mathit{NodeNumAcc}$): Binary indicator equal to 1 if the predicted node count matches the ground truth, and 0 otherwise.

\textbf{Connection Accuracy} ($\mathit{ConnectAcc}$): Measures topological correspondence using the Hungarian algorithm for node matching, evaluated separately for SE and VE connections through F1-score computation:
\begin{equation}
\mathit{ConnectAcc} = \frac{\mathit{SE\_ConnectAcc} + \mathit{VE\_ConnectAcc}}{2}
\end{equation}

\textbf{Topological Similarity} ($\mathit{TopologySimilarity}$): Compares Laplacian eigenvalues between predicted and ground truth graphs, where $\boldsymbol{\lambda}$ denotes the vector of sorted Laplacian eigenvalues:
\begin{equation}
\mathit{TopologySimilarity} = 1 - \frac{\|\boldsymbol{\lambda}_{\text{pred}} - \boldsymbol{\lambda}_{\text{true}}\|_2}{\max(\|\boldsymbol{\lambda}_{\text{pred}}\|_2, \|\boldsymbol{\lambda}_{\text{true}}\|_2)}
\end{equation}

\textbf{Position Accuracy} ($\mathit{PositionAcc}$): Measures the spatial alignment of matched node pairs using the Mean Squared Error (MSE) on normalized coordinates, converted to a bounded accuracy score:
\begin{equation}
\mathit{PositionAcc} = \frac{1}{1 + \text{MSE}(\bar{\mathbf{p}}_{\text{pred}},\, \bar{\mathbf{p}}_{\text{gt}})}
\end{equation}
where $\bar{\mathbf{p}}$ denotes the bounding-box normalized coordinates.

\textbf{Node Size Accuracy} ($\mathit{NodesizeAcc}$): Measures the accuracy of predicted node size parameters:
\begin{equation}
\mathit{NodesizeAcc} = 1 - \frac{\text{MAE}(s_{\text{pred}},\, s_{\text{gt}})}{s_{\max}}
\end{equation}
where $s_{\max} = \max(s_{\text{pred}}, s_{\text{gt}})$ is the maximum node size in the sample, ensuring normalization relative to the characteristic scale of each instance.

Rather than a traditional penalty, this structural reward acts as a dynamic weighting mechanism:
\begin{equation}
\begin{split}
\mathit{Accuracy} = \mathit{NodeNumAcc} \times &(0.3 \cdot \mathit{ConnectAcc} + 0.3 \cdot \mathit{TopologySimilarity} \\
+ &0.2 \cdot \mathit{PositionAcc} + 0.2 \cdot \mathit{NodesizeAcc})
\end{split}
\end{equation}

The S2MS-Loss uses $(1 - \mathit{Accuracy})$ to penalize poorly structured outputs by amplifying the base cross-entropy:
\begin{equation}
\mathit{L}_{\text{S2MS}} = \bigl(1 + \beta \cdot (1 - \mathit{Accuracy})\bigr) \cdot L_{\text{CE}}
\end{equation}

where $\beta$ controls the penalty strength. When $\mathit{Accuracy}=1$, the multiplier is $1$ and only $L_{\text{CE}}$ remains; when $\mathit{Accuracy}=0$, the multiplier becomes $(1+\beta)$, amplifying gradients for topologically invalid predictions.

The overall training procedure is summarized as follows:
\begin{enumerate}
\item \textbf{Forward pass:} The VLM generates a skeletal description $\hat{y}$ from input sketch $x$.
\item \textbf{Structural evaluation:} Decode $\hat{y}$ and the ground truth $y$ into graph representations. Compute $\mathit{NodeNumAcc}$, $\mathit{ConnectAcc}$, $\mathit{TopologySimilarity}$, $\mathit{PositionAcc}$, and $\mathit{NodesizeAcc}$ via graph matching.
\item \textbf{Accuracy aggregation:} $\mathit{Accuracy} = \mathit{NodeNumAcc} \times (0.3 \cdot \mathit{ConnectAcc} + 0.3 \cdot \mathit{TopologySimilarity} + 0.2 \cdot \mathit{PositionAcc} + 0.2 \cdot \mathit{NodesizeAcc})$.
\item \textbf{Loss computation:} Compute $L_{\text{CE}}$, then scale: $L_{\text{S2MS}} = (1 + \beta \cdot (1 - \mathit{Accuracy})) \cdot L_{\text{CE}}$.
\item \textbf{Backward pass:} Standard backpropagation through the VLM; the structural $\mathit{Accuracy}$ contributes as a scalar multiplier on the loss gradient.
\end{enumerate}

\subsubsection{Training strategy}

The complete ablation study is based on Qwen2.5-VL-72B-Instruct~\cite{wang2024qwen2vl,bai2025qwen25vl}, with additional architectures evaluated in Table~\ref{tab:multi_model_comparison}. We employ a two-stage fine-tuning strategy: stage one uses $\beta = 0.5$ to acquire structural generation capabilities with cross-entropy dominant; stage two uses $\beta = 1.6$ to amplify topological constraints. Training is conducted on 2$\times$NVIDIA RTX PRO 6000 GPUs using LLaMA-Factory and Unsloth~\cite{zheng2024llamafactory,han2023unsloth}.
\section{Results}

\subsection{Quantitative evaluation and ablation studies}

We systematically analyzed the influence of inference and training hyperparameters on model performance. The full ablation is provided in Table~\ref{tab:summary_results} in the Appendix.

Increasing $\mathit{temperature}$ and $\mathit{top\_p}$ improved the score from 0.356 to 0.455; the best configuration ($\mathit{temperature}=1.00$, $\mathit{top\_p}=0.95$) was adopted for subsequent experiments. Performance peaked at 600 training steps, while excessive training caused slight overfitting. Step-by-step prompts increased topological similarity by 7\%, halving the data requirement without compromising accuracy.

Multimodal integration with standardized coordinates and two-stage fine-tuning yielded a peak score of 0.677, an 83.7\% improvement over the baseline (0.356)---with strong generalization across line drawings and shadow-overlaid inputs. Under zero-shot settings, general-purpose VLMs (GPT-4.1, Claude Sonnet 4, Gemini 2.5 Pro, Qwen2.5-VL) failed to produce topologically valid outputs, confirming that both the dataset and structured encoding are essential.

\begin{table}[h]
\centering
\small
\caption{Comparison with domain-specific sketch-to-3D baselines. Topological Similarity (higher is better), Manifold status, and Editability.}
\label{tab:baseline_comparison}
\setlength{\tabcolsep}{4pt}
\begin{tabular}{lccc}
\toprule
Method & Topo. Sim. $\uparrow$ & Manifold & Editable \\
\midrule
Vitruvio~\cite{tono2024vitruvio} & 0.422 & no & no \\
BuilDiff~\cite{wei2023buildiff} & 0.287 & no & no \\
\textbf{Sketch2MinSurf (Ours)} & \textbf{0.844} & \textbf{yes} & \textbf{yes} \\
\bottomrule
\end{tabular}
\end{table}

We further compare against domain-specific sketch-to-3D baselines from the architectural design literature. As shown in Table~\ref{tab:baseline_comparison}, existing methods produce non-manifold geometries that cannot be directly edited or fabricated, whereas Sketch2MinSurf achieves substantially higher topological similarity while ensuring manifold output.

To evaluate the reward-modulated objective across architectures, we conducted comparative experiments (Table~\ref{tab:multi_model_comparison}).

\begin{table*}[htbp]
    \centering
    \caption{Effect of reward-modulated structural objective across different VLM architectures. The table compares the performance of five models under standard cross-entropy loss (Single-stage) and the proposed reward-modulated objective (Two-stage +Reward).}
    \label{tab:multi_model_comparison}
    \resizebox{\textwidth}{!}{
    \begin{tabular}{@{}l@{\hspace{10pt}}c@{\hspace{10pt}}c@{\hspace{10pt}}c@{\hspace{10pt}}c@{\hspace{10pt}}c@{\hspace{10pt}}c@{\hspace{10pt}}c@{}}
        \toprule
        \textbf{Model Architecture} &
        \makecell[c]{\textbf{Training}\\\textbf{Stage}} &
        \makecell[c]{\textbf{Training}\\\textbf{Steps}} &
        \makecell[c]{\textbf{Evaluation}\\\textbf{Score}} &
        \makecell[c]{\textbf{Connection}\\\textbf{Acc.}} &
        \makecell[c]{\textbf{Position}\\\textbf{Acc.}} &
        \makecell[c]{\textbf{Topological}\\\textbf{Sim.}} &
        \makecell[c]{\textbf{NodeNum}\\\textbf{Acc.}} \\
        \midrule
        Qwen2.5-VL-72B-Instruct & Single & 600 & 0.642 & 0.235 & 0.799 & 0.817 & 1.000 \\
        Qwen2.5-VL-72B-Instruct (\textbf{+Reward}) & Two-stage & 600 & \textbf{0.654} & \textbf{0.255} & \textbf{0.832} & \textbf{0.818} & \textbf{1.000} \\
        \midrule
        Qwen3-VL-32B-Instruct & Single & 600 & 0.653 & 0.241 & 0.843 & 0.822 & 1.000 \\
        Qwen3-VL-32B-Instruct (\textbf{+Reward}) & Two-stage & 600 & \textbf{0.672} & \textbf{0.255} & \textbf{0.865} & \textbf{0.840} & \textbf{1.000} \\
        \midrule
        Qwen3.5-27B & Single & 600 & 0.658 & 0.250 & 0.835 & 0.823 & 1.000 \\
        Qwen3.5-27B (\textbf{+Reward, best}) & Two-stage & 600 & \textbf{0.677} & \textbf{0.267} & \textbf{0.859} & \textbf{0.844} & \textbf{1.000} \\
        \midrule
        InternVL3-78B & Single & 600 & 0.596 & 0.179 & 0.835 & 0.701 & 1.000 \\
        InternVL3-78B (\textbf{+Reward}) & Two-stage & 600 & \textbf{0.645} & \textbf{0.234} & \textbf{0.840} & \textbf{0.800} & \textbf{1.000} \\
        \midrule
        Llama-3.2-90B-Vision & Single & 600 & 0.585 & 0.208 & 0.790 & 0.677 & 1.000 \\
        Llama-3.2-90B-Vision (\textbf{+Reward}) & Two-stage & 600 & \textbf{0.625} & \textbf{0.211} & \textbf{0.763} & \textbf{0.819} & \textbf{1.000} \\
        \bottomrule
    \end{tabular}
    }
\end{table*}

Across all architectures, the reward-modulated structural objective consistently improved evaluation scores compared to standard cross-entropy training. By penalizing topologically invalid predictions more heavily, the mechanism steered models toward higher connection accuracy and topological similarity. These results demonstrate that the proposed training objective is architecture-agnostic and provides complementary structural supervision beyond standard cross-entropy optimization.

\subsection{Qualitative comparison}

\begin{figure*}[htbp]
  \centering
   \includegraphics[width=0.8\linewidth]{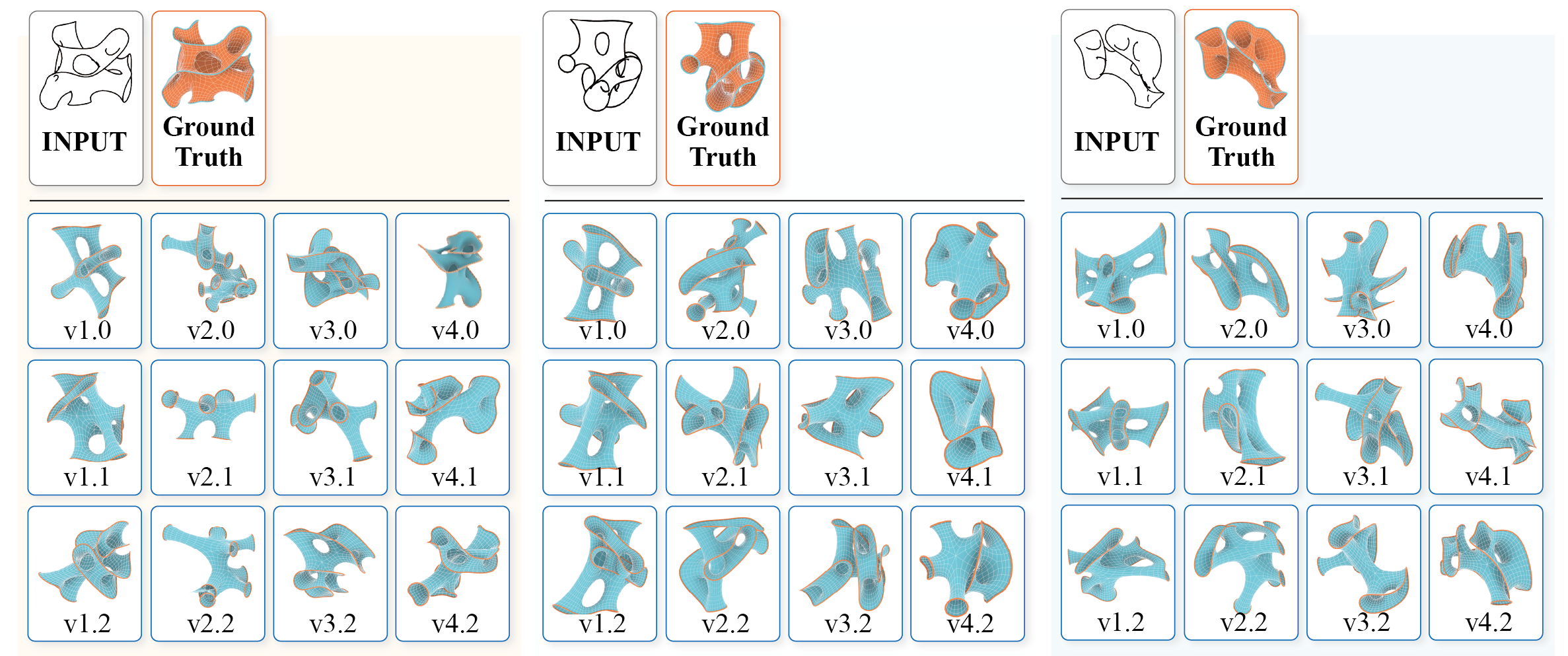}
   \caption{Comparative reconstruction results across model generations. The v4.2 series achieves the closest correspondence to ground truth.}
   \label{fig:comparative}
\end{figure*}

The reconstructed minimal surface skeletons, shown in Fig.~\ref{fig:comparative}, were generated using the Sketch2MinSurf Encoder and matched against the corresponding ground truth. Early models exhibited rotationally symmetric yet topologically redundant outputs, suggesting limited generalization. In the improved configurations, the introduction of the reward-modulated structural objective yielded higher structural continuity and geometric fidelity. The best model (v4.2) faithfully captured overall topology, void structure, and connectivity.

\begin{figure}[htbp]
  \centering
  \begin{minipage}[b]{0.5\linewidth}
    \centering
    \includegraphics[width=0.9\linewidth]{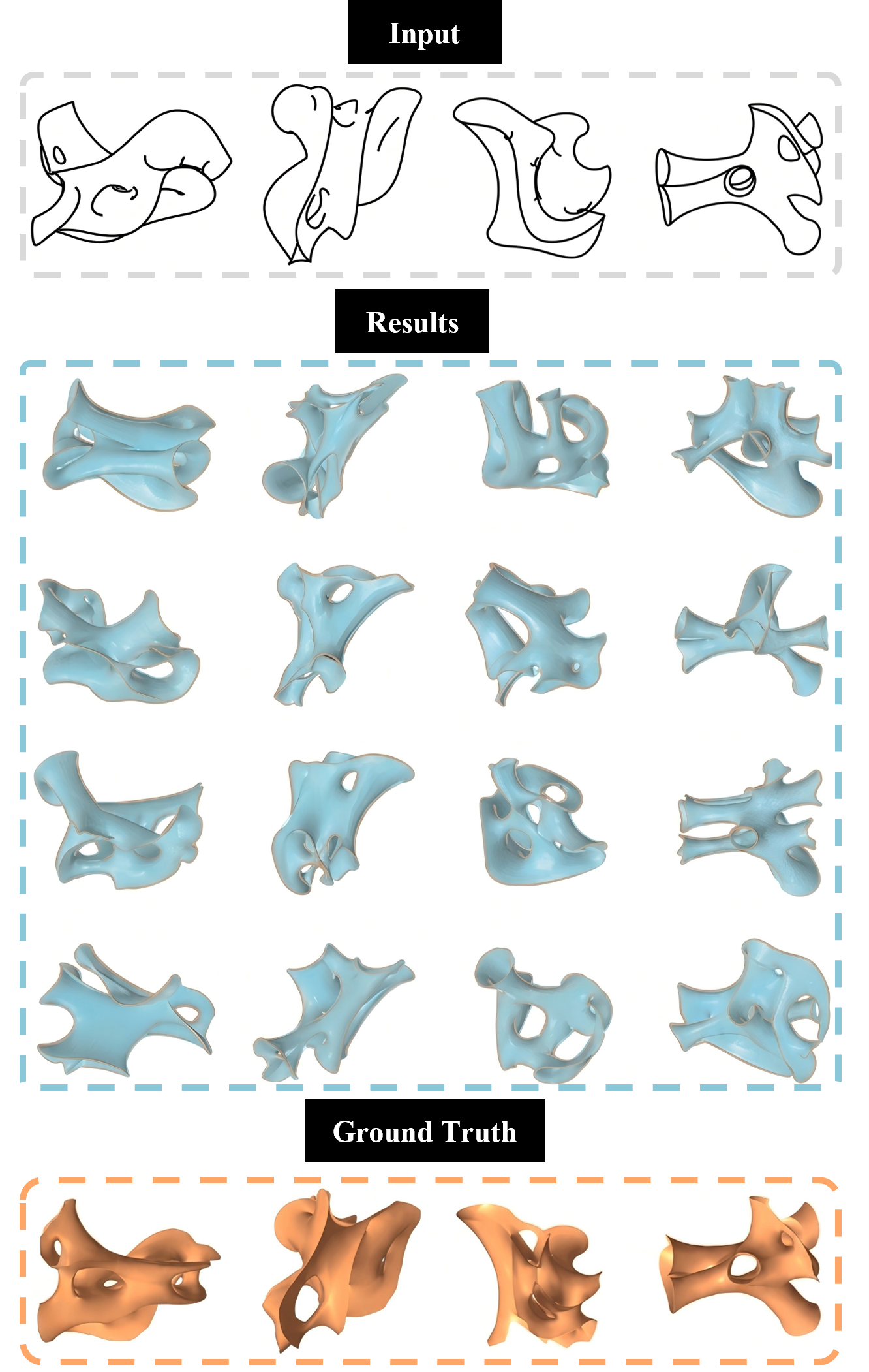}
  \end{minipage}
  \hfill
  \begin{minipage}[b]{0.44\linewidth}
    \centering
    \includegraphics[width=0.95\linewidth]{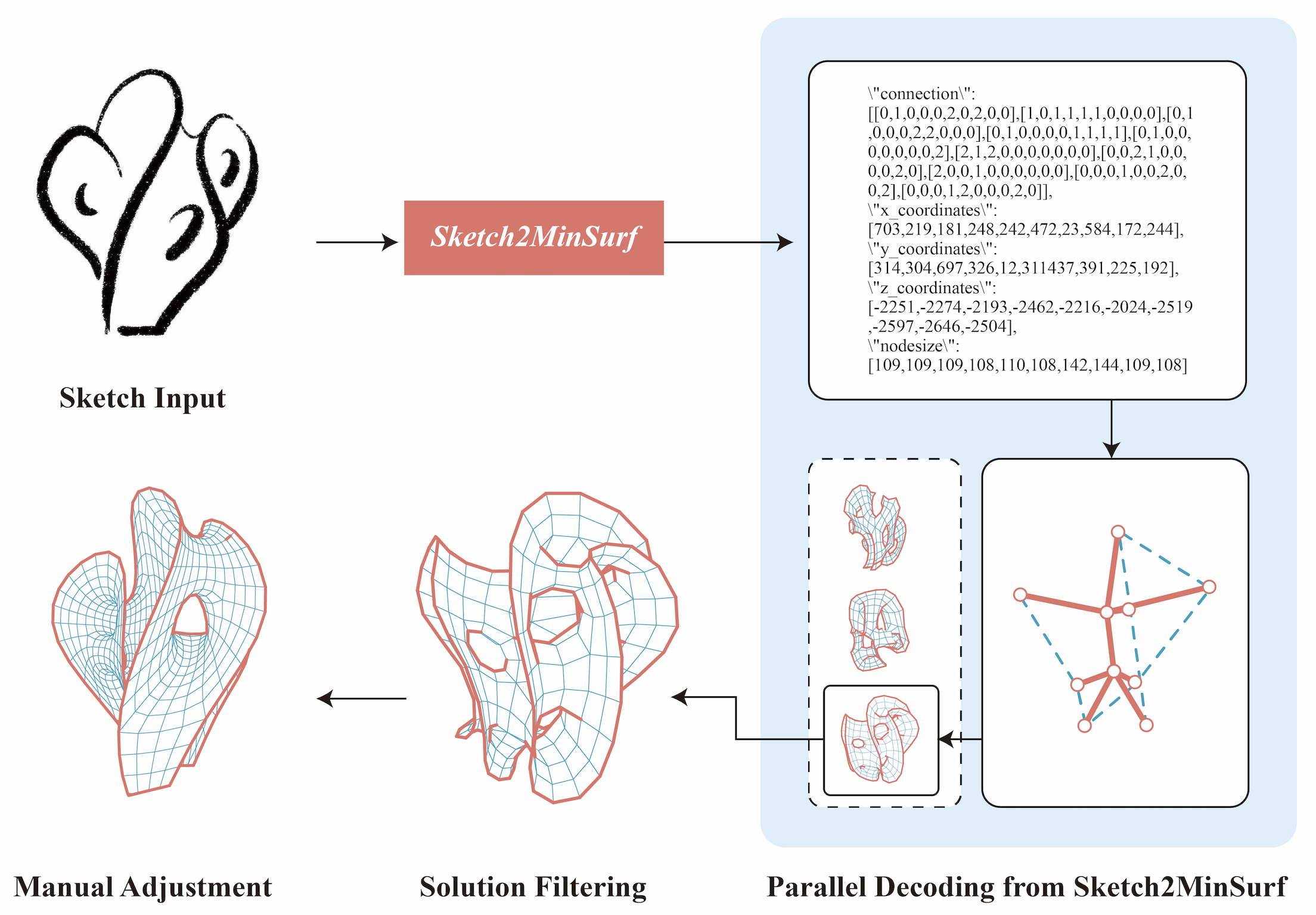}
    \vspace{0.15cm} 
    \includegraphics[width=0.95\linewidth]{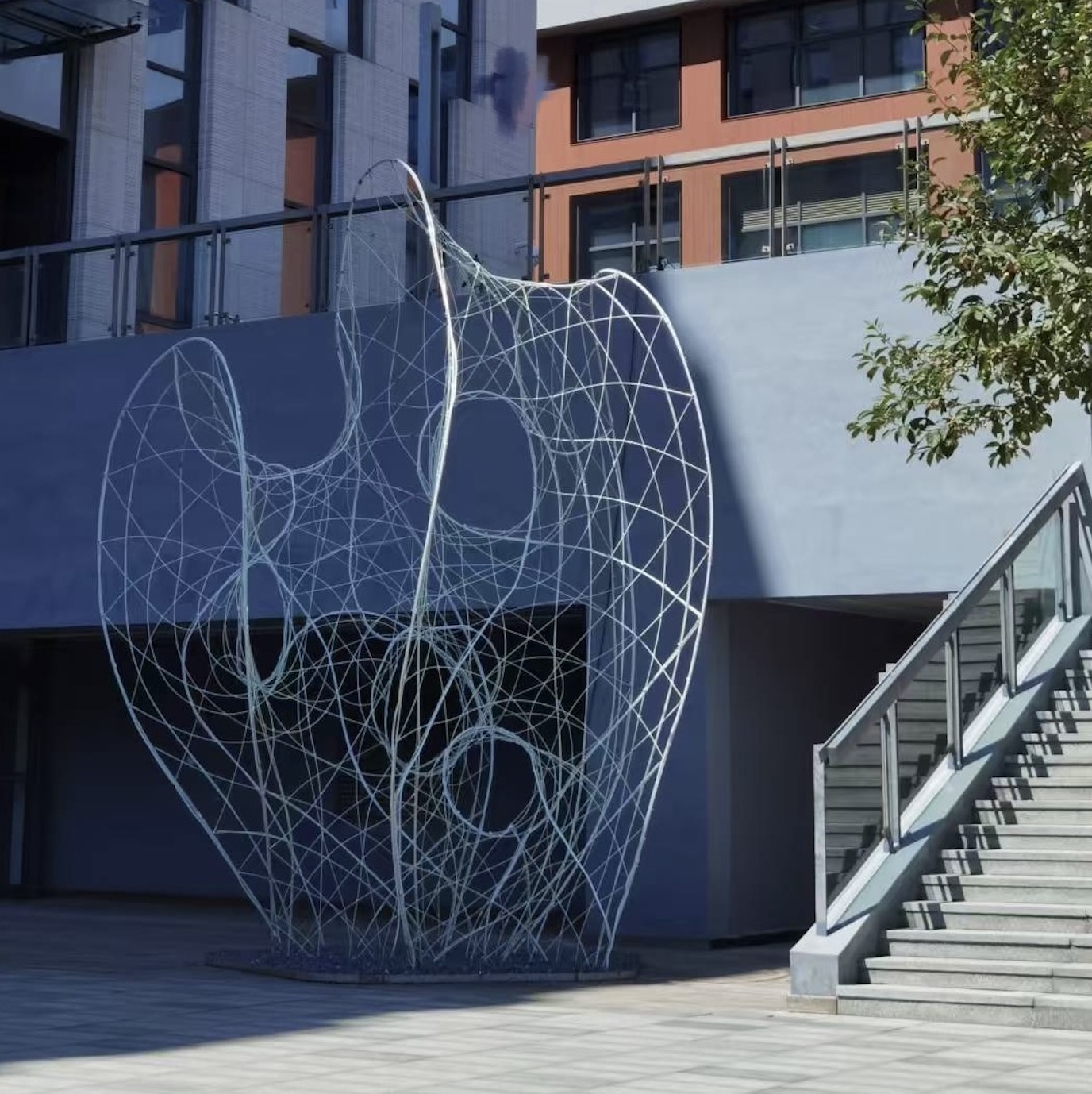}
  \end{minipage}
  
  \vspace{0.15cm} 
  \begin{minipage}[t]{0.54\linewidth}
    \caption{Performance of our best model (v4.2 series) on real hand-drawn sketches.}
    \label{fig:performance}
  \end{minipage}
  \hfill
  \begin{minipage}[t]{0.42\linewidth}
    \caption{Application: (top) design-to-construction workflow; (bottom) built installation (Aug 2025).}
    \label{fig:application}
  \end{minipage}
  
\end{figure}

Fig.~\ref{fig:performance} illustrates inference results from real hand-drawn sketches. Our best model reconstructs manifold structures that capture both local geometry and global topology, confirming that the learned representation generalizes from synthetic renderings to freehand input.

\subsection{Application in architectural practice}

The Sketch2MinSurf pipeline was deployed in a university design project to construct a full-scale minimal surface installation. Conceptual sketches were processed through the model to produce editable 3D topological skeletons, refined under site-specific constraints using dynamic relaxation in Grasshopper, and physically fabricated as a spatial gridshell structure with integrated lighting, merging architectural form-finding and computational precision. This demonstrates the system’s capacity to translate conceptual sketches into real constructed artifacts satisfying both design intent and structural logic. Additional details and photographs are provided in the supplementary material.

\section{Discussion}

The Sketch2MinSurf framework addresses the challenge of generating valid 3D manifold surfaces directly from single-view architectural sketches. By establishing a three-layer mapping across design intent, spatial reasoning, and geometric generation, it advances consistency, generality, and controllability in sketch-based modeling of complex non-Euclidean forms.

The Sketch2MinSurf Description encodes node coordinates, solid and void edges, and node sizes into a robust parametric representation. Unlike conventional boundary-based modeling~\cite{piegl2012nurbs}, this formulation operates on explicit connectivity, bridging visual sketch perception with topological reasoning and improving scalability to high-dimensional configurations. The S2MS-Loss reinforces this encoding through a reward-modulated mechanism: by dynamically scaling the cross-entropy loss according to structural accuracy, it guides the model toward topologically faithful outputs without requiring differentiable graph operations. Combined with the skeleton encoding, this loss enables direct translation to smooth minimal surfaces compatible with parametric design environments such as Grasshopper. By decoupling the VLM prediction from the geometric solver, the pipeline maintains full geometric editability and enforces fabrication constraints (watertightness, curvature continuity) that purely neural representations cannot guarantee, distinguishing it from voxel or NeRF-based pipelines~\cite{kim2020citycraft,tono2024vitruvio,asmar2020machinic}.

At the spatial reasoning level, Sketch2MinSurf integrates cognitive understanding of topological connectivity rather than relying solely on pixel- or point-cloud-based representations~\cite{luo2021diffusion3dpoint,yang2019pointflow,lin2018efficientpointcloud}. The system formalizes architectural sketch logic---expressive continuity of solids and voids---into interpretable skeleton operations. Even for line drawings without shadow cues, our best model consistently achieves topological similarity above 0.84, reflecting strong generalization for complex spatial configurations.

\subsection{Limitations and future work}

Despite these results, several limitations remain. When modeling topologies exceeding 16 nodes, the model exhibits reduced generalization; future work will incorporate graph neural networks to capture hierarchical relationships. The current single-view input constrains depth perception; integrating multi-view sketches could enhance spatial completeness. Future integration of structural stress and wall thickness via real-time simulation (e.g., Kangaroo) will support structural optimization and improve constructability.

\section{Conclusion}

We presented \textbf{Sketch2MinSurf}, a topology-driven framework that transforms hand-drawn sketches into editable, manifold 3D minimal surfaces. Our approach encodes surface structures through node coordinates, edge typologies, and node scales, converting nonlinear geometric forms into structured data supported by a 3000-sample multimodal dataset developed through four iterative refinements.

A central contribution is the S2MS-Loss, a reward-modulated objective that evaluates skeleton quality across five dimensions and dynamically scales the cross-entropy loss to prioritize structurally valid outputs. Experiments across five VLM architectures confirm that this mechanism consistently improves topological fidelity.

Departing from GAN, NeRF, or diffusion-based frameworks, Sketch2MinSurf encodes minimal surface properties as structured language, enabling VLMs to model continuity, hierarchy, and the interplay of solid and void. Our results show that formalizing architectural topology within a learning framework expands AI's spatial reasoning and opens new directions for human--AI collaboration in computational design. Overall, this work bridges design theory and computer vision, positioning topological encoding as a foundation for interpretable and generative 3D spatial modeling.

\medskip

{
\small
\bibliographystyle{ieeenat_fullname}
\bibliography{arxiv}
}

\appendix
\clearpage
\setcounter{page}{1}
\appendix

\section{Extended Related Work}

Overall, prior research has advanced image-to-3D reconstruction, multimodal spatial reasoning, and minimal-surface generation across three complementary directions. However, existing approaches have yet to jointly interpret design intent, ensure topological consistency, and maintain geometric editability. Our proposed framework addresses this gap by integrating VLM-driven spatial understanding with parametric geometry modeling, enabling scalable and semantically aligned minimal-surface generation from hand-drawn sketches.

The skeleton-based approach underlying Sketch2MinSurf translates directly into parametric generative workflows while maintaining computational efficiency for large-scale models. By decoupling the VLM prediction stage from the geometric solving stage, the pipeline preserves full editability and enforces fabrication constraints that purely neural implicit representations cannot guarantee.

\section{Extended Background on Minimal Surfaces in Architecture}

Architectural and structural design has long embraced minimal and near-minimal surfaces for their capacity to combine expressive form with material efficiency (Table~\ref{tab:minimal_surface_applications}). Built examples like the \textit{Metropol Parasol}, the \textit{ICD/ITKE Research Pavilion 2015--16}, the \textit{Agent Crystalline Pavilion}, and Toyo Ito's \textit{Taichung Metropolitan Opera House} demonstrate large-span, curvature-controlled shells derived from computational modeling and digital fabrication.

\begin{table*}[htbp]
    \centering
    \caption{Applications of minimal surfaces and similar surfaces across multiple scales of architectural studies.}
    \label{tab:minimal_surface_applications}
    \resizebox{\textwidth}{!}{
    \begin{tabular}{
        m{1.2cm}
        m{1.6cm}
        m{1.6cm}
        m{0.7cm}
        m{1.6cm}
        m{3.0cm}
        m{2.8cm}
    }
        \toprule
        \textbf{Scale} &
        \textbf{Project} &
        \textbf{Designer} &
        \textbf{Year} &
        \textbf{Location} &
        \textbf{Structural System} &
        \textbf{Photo} \\
        \midrule

        Urban or Pavilion &
        Metropol Parasol &
        J. Mayer H. Architects &
        2011 &
        Seville, Spain &
        Large timber grid-shell on concrete cores &
        \includegraphics[width=2.8cm, height=1.6cm]{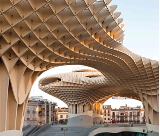} \\

        \midrule
        Urban or Pavilion &
        Agent Crystalline Pavilion &
        MARC FORNES / THEVERYMANY &
        2019 &
        Edmonton, Canada &
        Self-supporting laminated aluminum folded-plate shell &
        \includegraphics[width=2.8cm, height=1.6cm]{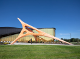} \\

        \midrule
        Building &
        Heydar Aliyev Center &
        Zaha Hadid Architects &
        2012 &
        Baku, Azerbaijan &
        Steel space-frame with curved secondary framing; composite cladding &
        \includegraphics[width=2.8cm, height=1.6cm]{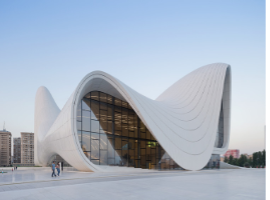} \\

        \midrule
        Building &
        National Taichung Theater &
        Toyo Ito \& Associates, Architects &
        2014 &
        Taichung, Taiwan &
        ``Sound Cave'' continuous reinforced-concrete shell/wall system &
        \includegraphics[width=2.8cm, height=1.6cm]{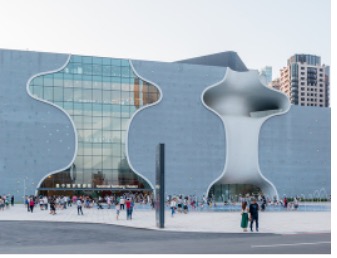} \\

        \midrule
        Building &
        Harbin Opera House &
        MAD Architects &
        2015 &
        Harbin, China &
        Steel space-frame shells and folded roof plates &
        \includegraphics[width=2.8cm, height=1.6cm]{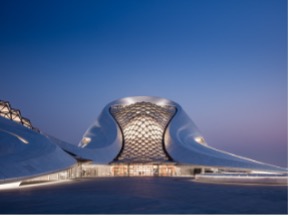} \\

        \midrule
        \makecell[l]{Structural} &
        Armadillo Vault &
        Block Research Group (ETH Z\"urich) &
        2016 &
        Venice, Italy &
        Unreinforced compression-only stone vault (dry assembly) &
        \includegraphics[width=2.8cm, height=1.6cm]{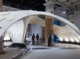} \\

        \midrule
        \makecell[l]{Structural} &
        3D Woven Chinese Knot &
        WX Studio &
        2017 &
        Beijing, China &
        Spatial lattice/tensegrity woven into a knot-like curved volume &
        \includegraphics[width=2.8cm, height=1.6cm]{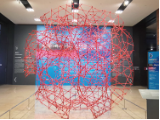} \\

        \midrule
        \makecell[l]{Structural} &
        KnitCandela &
        BRG + ZHA CODE + UNAM &
        2018 &
        Mexico City, Mexico &
        Thin concrete shell cast on knitted stay-in-place textile formwork &
        \includegraphics[width=2.8cm, height=1.6cm]{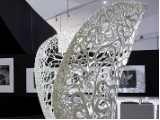} \\

        \midrule
        \makecell[l]{Structural} &
        The Orb &
        MARC FORNES / THEVERYMANY &
        2018 &
        El Paso, Texas, USA &
        Doubly-curved strip-shell with self-supporting monocoque action &
        \includegraphics[width=2.8cm, height=1.6cm]{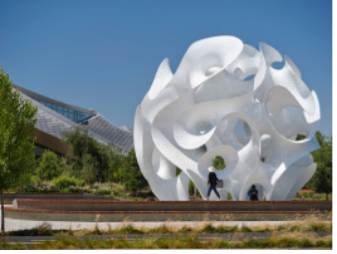} \\

        \midrule
        \makecell[l]{Structural} &
        BUGA Fibre Pavilion &
        ICD/ITKE University of Stuttgart &
        2019 &
        Heilbronn, Germany &
        Robotically wound fiber-composite shells on reusable formwork &
        \includegraphics[width=2.8cm, height=1.6cm]{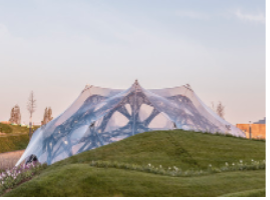} \\

        \bottomrule
    \end{tabular}
    }
\end{table*}

Contemporary design groups such as THEVERYMANY and the Tsinghua University Bending-Active Structures Lab experiment with minimal-surface lattices enriched by cultural ornamentation, pushing fabrication limits through robotic weaving and bespoke material exploration. These precedents highlight the architectural viability of the geometries our system generates.

\section{Extended Method Details}

\subsection{Mathematical and parametric foundations}

Minimal surfaces are formally defined as surfaces with zero mean curvature $H=0$ everywhere, satisfying the Euler--Lagrange equation for surface area minimization. Building upon Douglas's solution~\cite{douglas1939minimalsurfaces}, Sadoc and Charvolin~\cite{sadoc1989infinite} introduced graph-theoretic representations enabling topological classification of triply periodic minimal surfaces (TPMS).

Further computational models have incorporated energy-based optimization~\cite{wang2009minkowski} and polynomial approximation~\cite{andersson1988minimalsurfaces} to parameterize surface patches. Recent work~\cite{oka2015transformation,chen2020liquidframeworks,xiang2023blockcopolymer} enables curvature tuning and morphological transition control.

Our approach draws on these principles but embeds them within a differentiable learning framework, allowing the model to produce \textit{editable, geometry-regularized surfaces} directly from sketch input---bridging parametric modeling with data-driven reasoning.

\subsection{Binary unit description framework}

\begin{figure}[htbp]
  \centering
   \includegraphics[width=1.0\linewidth]{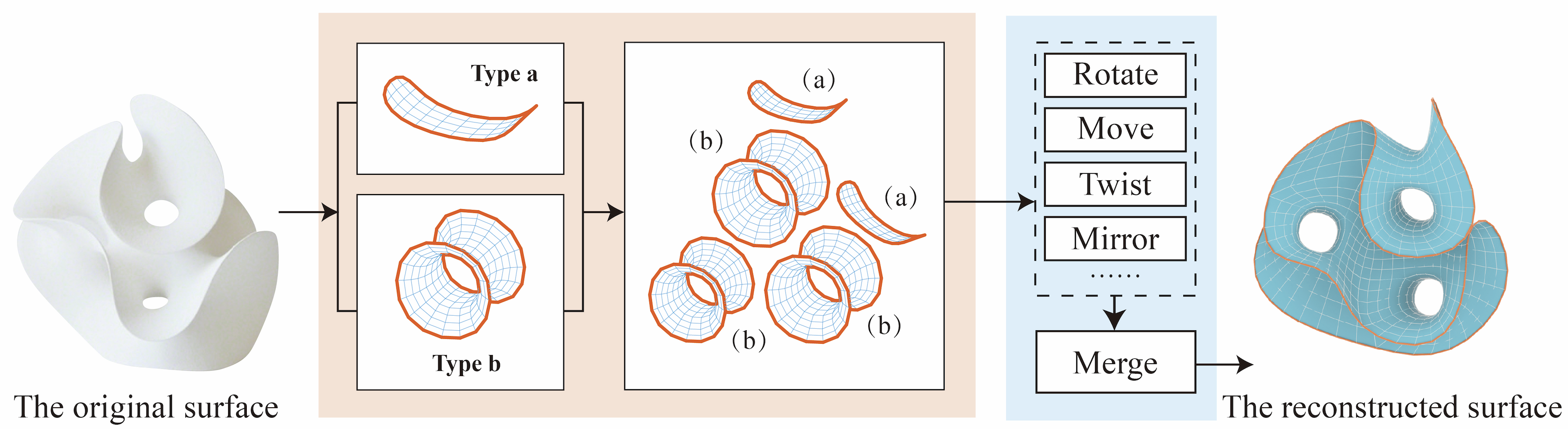}
   \caption{Framework of the minimal surface binary unit description method.}
   \label{fig:binary_unit_description_method_framework}
\end{figure}

\begin{figure}[htbp]
  \centering
   \includegraphics[width=1.0\linewidth]{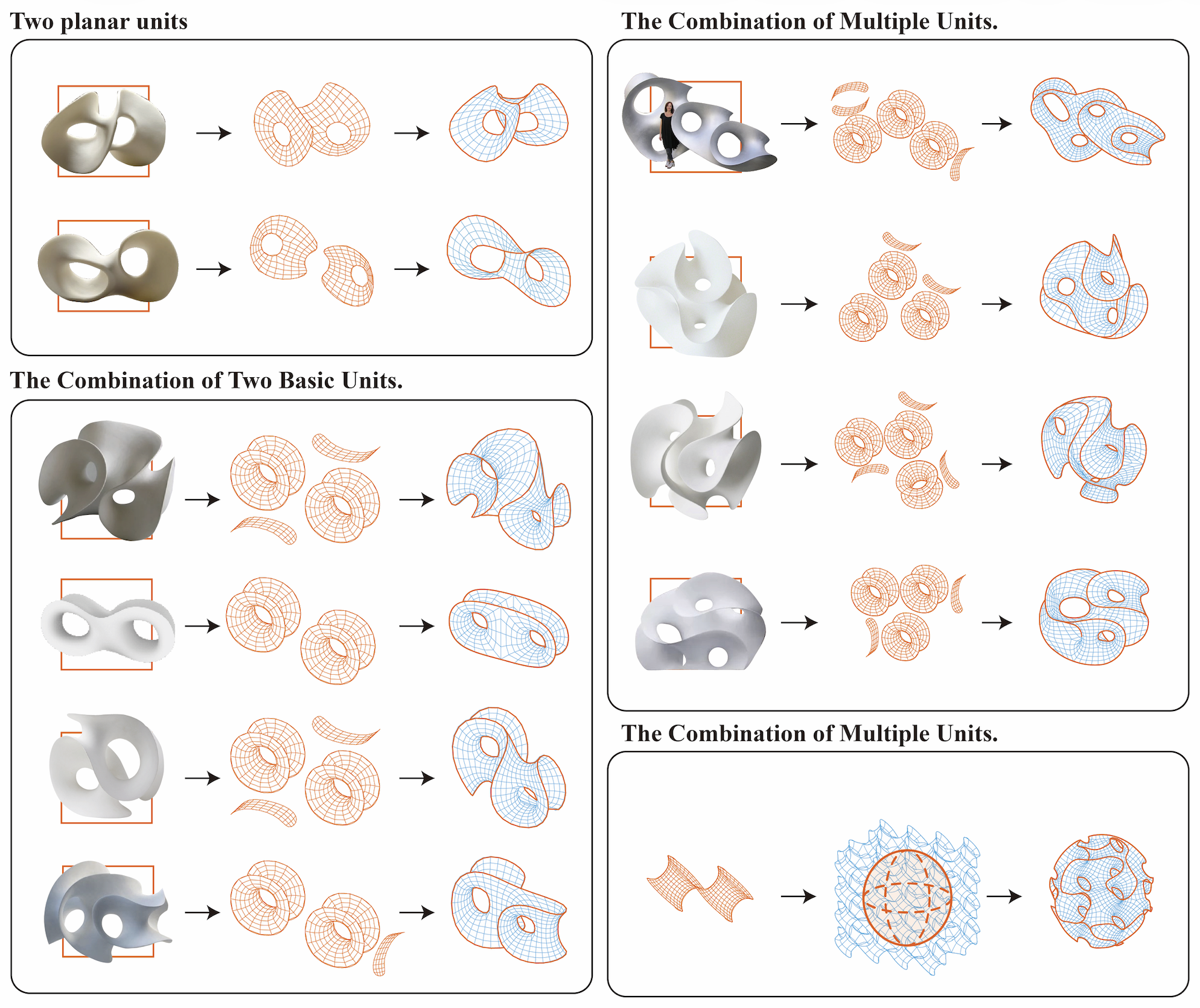}
   \caption{Schematic diagram of the minimal surface binary unit description method.}
   \label{fig:binary_unit_description_method_schematic}
\end{figure}

The binary unit approach (Fig.~\ref{fig:binary_unit_description_method_framework}) demonstrates that minimal surfaces can be described through the composition of Chen--Gackstatter surfaces with tubular surface elements. The Chen--Gackstatter surface decomposes into: (a) saddle regions defining negative curvature as saddle surface units, and (b) cylindrical surfaces inheriting triple rotational symmetry from the Enneper surface. Through parametric adjustment and geometric group operations (rotation, translation, twisting), this method generates diverse spatial structures including porous networks, dendritic fractals, and aperiodic tessellations (Fig.~\ref{fig:binary_unit_description_method_schematic}).

\subsection{Sketch2MinSurf encoder implementation}

\begin{figure}[htbp]
  \centering
   \includegraphics[width=1.0\linewidth]{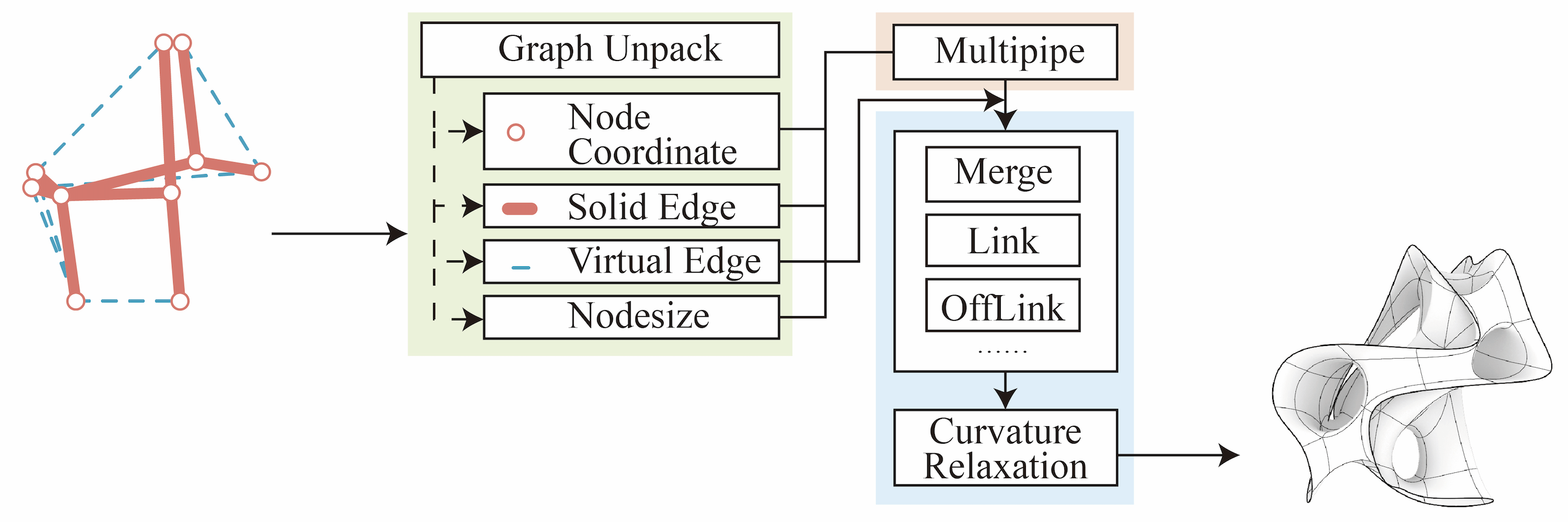}
   \caption{Detailed framework of the Sketch2MinSurf encoder.}
   \label{fig:Sketch2MinSurf_encoder}
\end{figure}

The complete encoding pipeline (Fig.~\ref{fig:Sketch2MinSurf_encoder}) begins by unpacking the graph representation, where the node coordinates define the spatial framework and the solid-edge (SE) connectivity specifies the primary structural skeleton. Virtual edges (VE) are then incorporated to introduce auxiliary connections that guide subsequent topology construction. This enriched graph is passed to the multi-tube module, which generates edge-aligned geometric primitives along both SE and VE connections. Within this module, a sequence of merge, link, and offset operations is applied to ensure that the resulting primitives form a watertight and topologically consistent surface. Finally, a curvature-relaxation stage iteratively smooths the geometry, effectively reducing mean curvature while preserving the encoded structure, and produces the final surface output.

\subsection{Virtual edge and solid edge skeleton derivation}

\begin{figure}[htbp]
  \centering
   \includegraphics[width=0.8\linewidth]{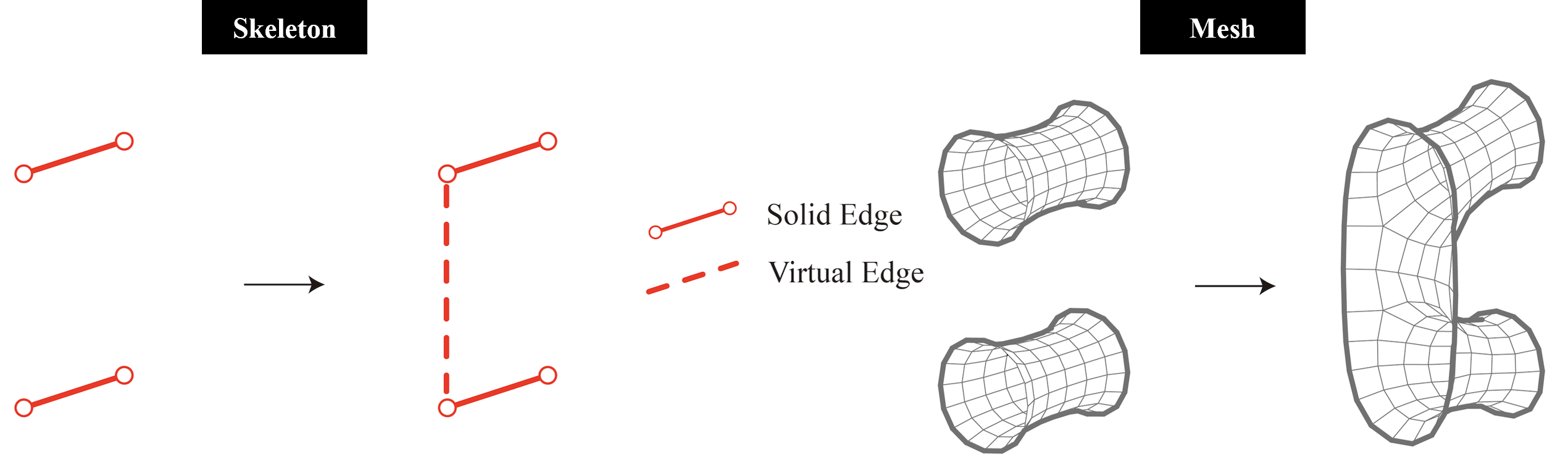}
   \caption{Illustration of virtual edge skeleton and solid edge skeleton derivation.}
   \label{fig:illustration_supp}
\end{figure}

Fig.~\ref{fig:illustration_supp} illustrates how the two edge types are derived from the surface topology. Solid edges trace the primary pipe centerlines and define where saddle patches meet, forming the hard boundary skeleton. Virtual edges connect nodes perpendicular to the solid-edge directions, guiding tubular transitions between saddle regions without imposing hard boundaries. Three connection operators (Link, Merge, and OffLink) determine how virtual-edge tubes join the solid-edge surface, enabling diverse topological configurations such as branching, merging, and offset connections.

\subsection{Detailed loss component formulations}

\textbf{Cross-Entropy Loss} ($L_{\text{CE}}$): The foundation of S2MS-Loss is the standard cross-entropy loss for instruction following:
\begin{equation}
L_{\text{CE}} = -\frac{1}{m} \sum_{i=1}^{m} \sum_{j=1}^{n} y_{ij} \log(a_{ij})
\end{equation}
where $m$ represents the number of samples, $n$ represents the number of categories, $y_{ij}$ is the one-hot encoded ground truth, and $a_{ij}$ is the predicted probability after softmax normalization.

\textbf{Connection Accuracy} ($\mathit{ConnectAcc}$): We focus on the agreement between the predicted and ground-truth graph connectivity, considering both the number of connections and their distribution across the graph. We first establish a one-to-one node correspondence using the Hungarian algorithm, pairing each predicted node with its closest counterpart in the ground truth. Connectivity is evaluated separately for solid edges ($\mathit{SE\_ConnectAcc}$) and virtual edges ($\mathit{VE\_ConnectAcc}$) through F1-score computation:
\begin{equation}
\mathit{Precision} = \frac{\mathit{TP}}{\mathit{TP} + \mathit{FP}}, \quad
\mathit{Recall} = \frac{\mathit{TP}}{\mathit{TP} + \mathit{FN}}
\end{equation}
\begin{equation}
\mathit{SE\_ConnectAcc} = \frac{2 \cdot \mathit{Precision} \cdot \mathit{Recall}}
{\mathit{Precision} + \mathit{Recall}}
\end{equation}
where $\mathit{TP}$ denotes correctly predicted edges, $\mathit{FP}$ denotes predicted but non-existent edges, and $\mathit{FN}$ denotes missed edges. The score for virtual edges, $\mathit{VE\_ConnectAcc}$, is computed identically.

\textbf{Position Accuracy} ($\mathit{PositionAcc}$): After establishing node correspondence, we compute the MSE on bounding-box normalized coordinates:
\begin{equation}
\mathit{PositionErr} = \frac{1}{n} \sum_{i=1}^{n} \|\bar{\mathbf{p}}_i^{\text{pred}} - \bar{\mathbf{p}}_i^{\text{true}}\|_2^2, \quad
\mathit{PositionAcc} = \frac{1}{1 + \mathit{PositionErr}}
\end{equation}

\textbf{Node Size Accuracy} ($\mathit{NodesizeAcc}$): Let $s_i^{\text{pred}}$ and $s_i^{\text{gt}}$ denote the predicted and ground-truth size parameter of the $i$-th node. We compute the Mean Absolute Error and normalize by $s_{\max} = \max(s_{\text{pred}}, s_{\text{gt}})$:
\begin{equation}
\text{MAE} = \frac{1}{n} \sum_{i=1}^{n} |s_i^{\text{pred}} - s_i^{\text{gt}}|, \quad
\mathit{NodesizeAcc} = \max\left(0, 1 - \frac{\text{MAE}}{s_{\max}}\right)
\end{equation}
This ensures $\mathit{NodesizeAcc} \in [0, 1]$, where values closer to $1$ correspond to more accurate predictions.

\section{Dataset Construction and Training Details}

\subsection{Dataset overview}

\begin{figure}[htbp]
  \centering
   \includegraphics[width=0.5\linewidth]{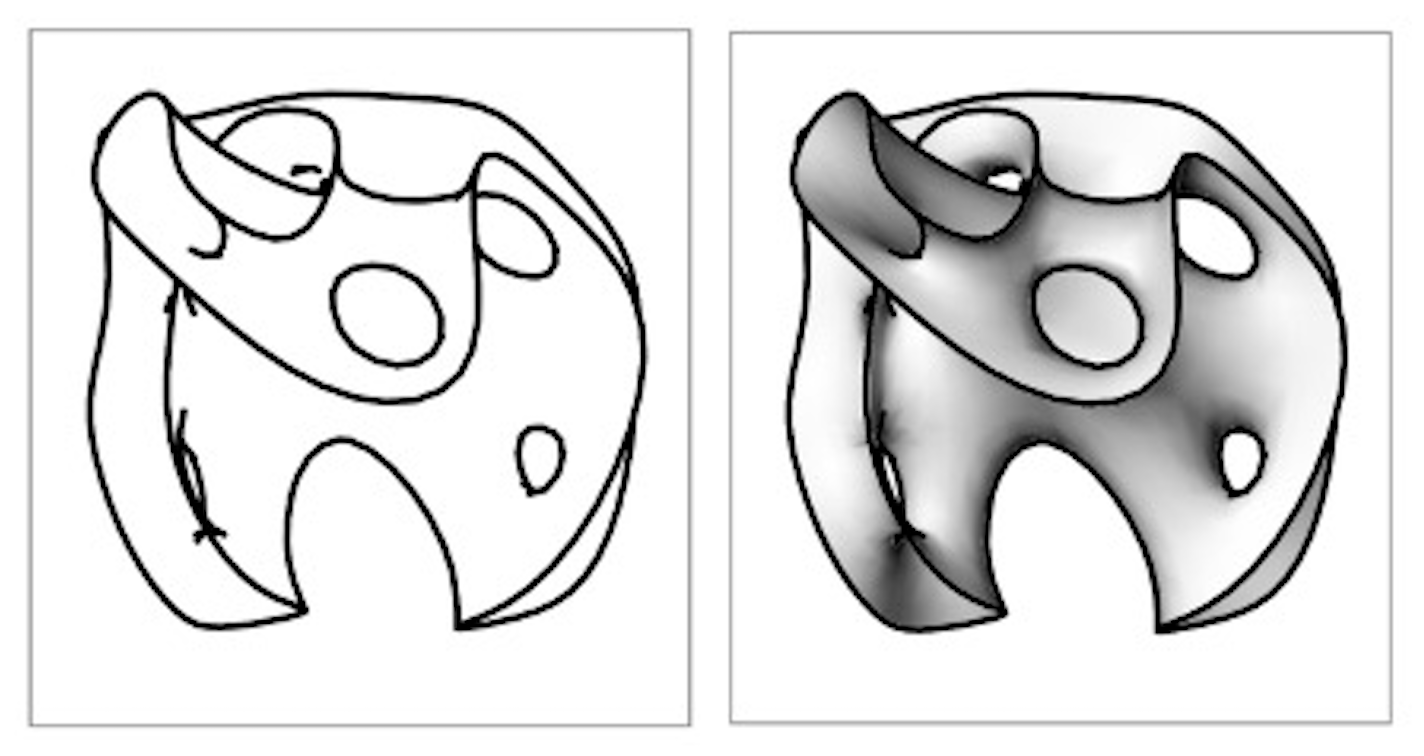}
   \caption{Image input example (line sketch on left, grayscale shade rendering overlaid with line sketch on right).}
   \label{fig:example_supp}
\end{figure}

\begin{figure}[htbp]
  \centering
   \includegraphics[width=0.9\linewidth]{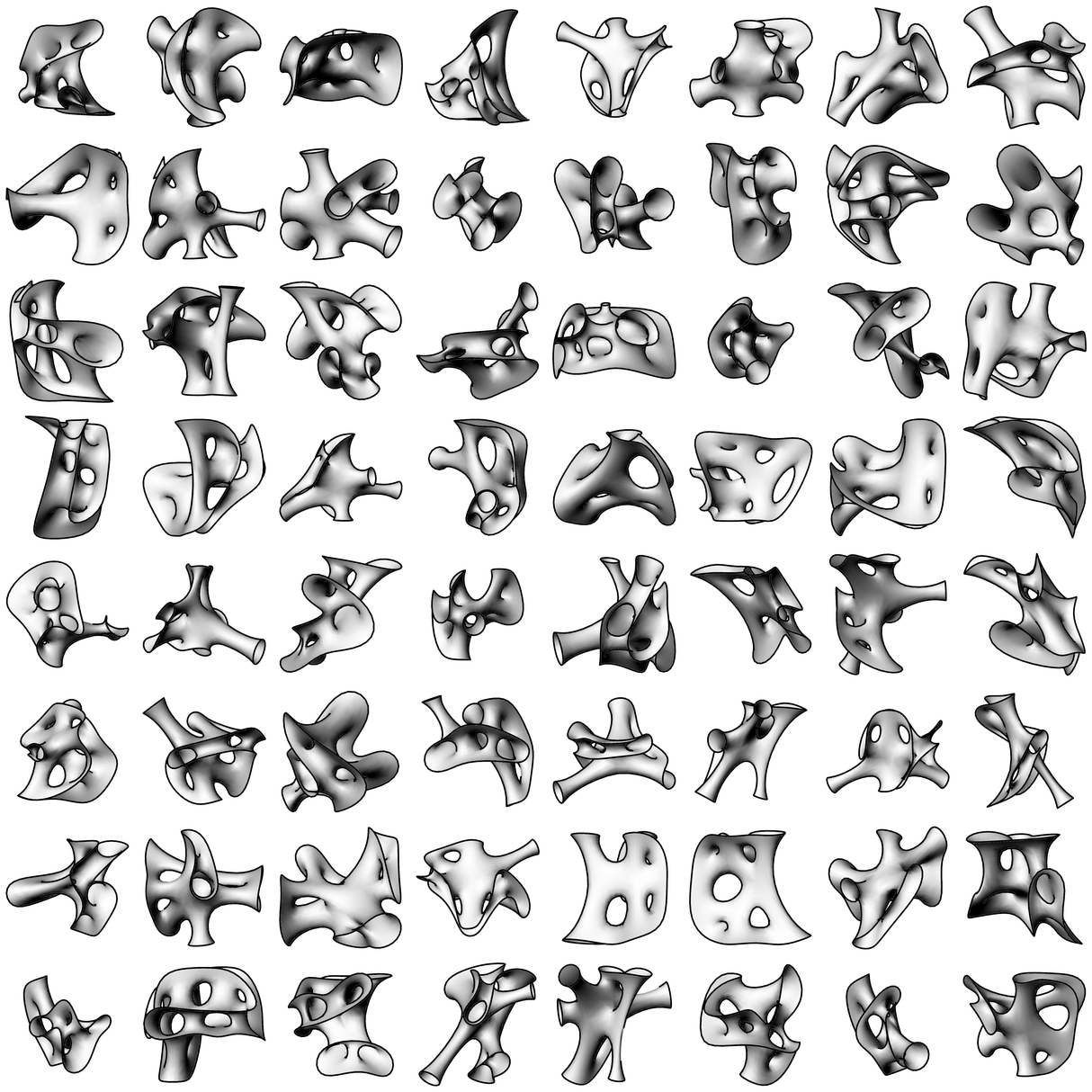}
   \caption{Overview of the training dataset. Each sample consists of a single-view rendering (750$\times$750) combining line drawing with grayscale shading, paired with a structured skeletal description.}
   \label{fig:dataset_overview}
\end{figure}

We construct a training dataset of 3{,}000 samples in OpenAI ShareGPT format with image-prompt pairs as input and structured skeletal descriptions as output. Input images are single-view renderings at 750$\times$750 resolution combining line drawings with grayscale shading (Fig.~\ref{fig:example_supp}). Fig.~\ref{fig:dataset_overview} shows an overview of representative samples from the dataset. The dataset was developed through four iterations (v1.0--v4.0), progressively refining encoding schemes, prompt designs, and visual representations to improve structural annotation quality and model training effectiveness.

\subsection{Prompt design}

We design three prompt variants across dataset iterations to evaluate the impact of task instruction granularity on model performance (Table~\ref{tab:prompt_variants}).

The \textbf{TextSimple} variant (v1) provides a concise task instruction: ``Please analyze this image and infer its possible topological structure. Output your answer in matrix form, including the adjacency matrix for connectivity, and the x, y, and z coordinate matrices.''

The \textbf{TextExact} variant (v2) adds a precise input description (``A line drawing representing the 2D planar projection of a contour, overlaid with shading, depicting an approximate minimal surface 3D model'') together with explicit output specifications: node count as an integer, connectivity as an adjacency matrix with elements 0 (no connection), 1 (solid edge), or 2 (void edge), and coordinate matrices in 2D array format using a relative coordinate system with the top-left node as origin.

The \textbf{TextDetailed} variant (v3/v4) extends TextExact with step-by-step reasoning guidance and switches to a camera-based coordinate system (x horizontal right, y vertical up, z perpendicular outward from the image plane). The inference method instructs the model to: (1) perform region segmentation to identify surface patches and voids, (2) extract skeletons from main regions where closed skeletons correspond to void edges and linear skeletons have endpoints as nodes, (3) detect internal lines not involved in region segmentation as additional nodes with void-edge connections, and (4) ensure full skeleton connectivity by extending nodes along the z-axis.

\begin{table}[htbp]
    \centering
    \caption{Summary of prompt variants designed for skeleton description generation.}
    \label{tab:prompt_variants}
    \setlength{\tabcolsep}{4pt}
    \begin{tabular}{@{}lp{9cm}@{}}
        \toprule
        \textbf{Variant} & \textbf{Description} \\
        \midrule
        TextSimple (v1) & Brief task instruction asking the model to analyze the image and infer the topological skeleton structure. \\
        \midrule
        TextExact (v2) & Precise input description, task objective, and output specifications: node count, adjacency matrix (0/1/2), and relative coordinate matrices. \\
        \midrule
        TextDetailed (v3/v4) & Extended prompt with step-by-step reasoning guidance (region segmentation, skeleton extraction, internal line detection, connectivity analysis) using camera-based coordinates. \\
        \bottomrule
    \end{tabular}
\end{table}

\subsection{Coordinate representation}

We evaluate two coordinate representation strategies for encoding node positions in the skeletal descriptions.

\textbf{Relative coordinate system:} Uses the upper-left node as the origin reference. All node coordinates are expressed as offsets relative to this anchor point, providing translation-invariant representations.

\begin{figure}[htbp]
  \centering
   \includegraphics[width=0.6\linewidth]{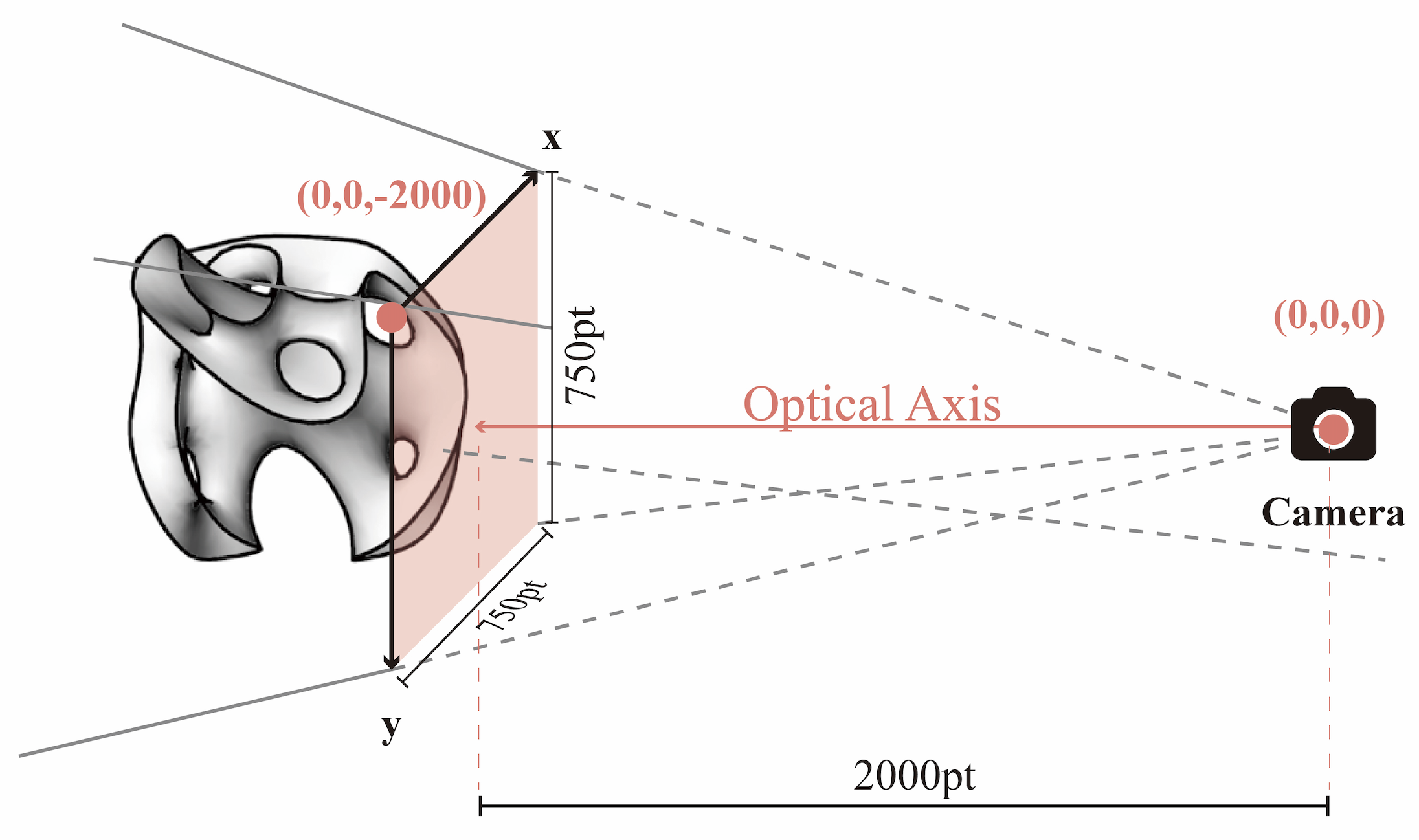}
   \caption{Camera-based coordinate representation method.}
   \label{fig:camera_supp}
\end{figure}

\textbf{Camera-based coordinate system:} Aligns coordinates with the camera focal plane, where $x$ and $y$ correspond to the image plane axes and $z$ corresponds to the depth axis (Fig.~\ref{fig:camera_supp}). This representation preserves the spatial relationship between the sketch view and the 3D structure, facilitating the model's understanding of depth from single-view images.

\subsection{Training hyperparameters}

All models are fine-tuned using LLaMA-Factory with Unsloth acceleration on 2$\times$NVIDIA RTX PRO 6000 GPUs. The base model is Qwen2.5-VL-72B-Instruct~\cite{wang2024qwen2vl,bai2023qwenvl,bai2025qwen25vl}. Key hyperparameters include: learning rate $2 \times 10^{-5}$ with cosine scheduling, batch size 1 with gradient accumulation over 8 steps, LoRA rank 64 and alpha 128 targeting all linear layers, and maximum sequence length 4{,}096 tokens. Inference uses temperature $T=1.0$ and top-$p=0.95$. Total training time for the two-stage pipeline is approximately 11 hours 40 minutes.

\subsection{Two-stage training strategy}

We employ a two-stage fine-tuning strategy. Stage one uses $\beta = 0.5$ with cross-entropy as the dominant objective, enabling the model to acquire basic structural generation capabilities. Stage two increases $\beta$ to 1.6 to amplify topological constraints, reinforcing the model's sensitivity to connectivity and spatial accuracy. This progressive scheduling prevents early training instability while ensuring strong topological reasoning in the final model.

\subsection{Evaluation methodology}

Model performance is evaluated using five held-out test sets of 100 images each, varying in visual processing and prompt design while maintaining identical subjects. For each test sample, the model generates a skeletal description from the input sketch, which is then decoded into a graph representation and compared against the ground truth across all five accuracy dimensions. The final Evaluation Score aggregates these metrics through the weighted sum:
\begin{equation}
\begin{split}
\mathit{Score} = \mathit{NodeNumAcc} \times &(0.3 \cdot \mathit{ConnectAcc} + 0.3 \cdot \mathit{TopologySimilarity} \\
+ &0.2 \cdot \mathit{PositionAcc} + 0.2 \cdot \mathit{NodesizeAcc})
\end{split}
\end{equation}
This metric is computed independently of the training loss, avoiding circular evaluation: during training, Accuracy serves as a reward-shaped scaling factor on cross-entropy, while at evaluation time it is computed directly against ground-truth skeletons to measure structural fidelity.

\section{Full Ablation Study}

Table~\ref{tab:summary_results} presents the comprehensive ablation study across all model configurations. The analysis reveals several key findings:

\textbf{Sampling diversity (v1.1):} Increasing temperature from 0.1 to 1.0 and top-$p$ from 0.5 to 0.95 yielded the largest single improvement in position accuracy (0.321$\to$0.712), confirming that diverse sampling helps the model explore the structured output space more effectively.

\textbf{Extended training (v1.2):} Training for 900 steps degraded performance across all metrics compared to 600 steps, indicating overfitting. Early stopping at 300 steps (v2.0) also underperformed, suggesting 600 steps as the optimal training duration.

\textbf{Step-by-step prompts (v2.1, v3.0):} The TextDetailed prompt improved topological similarity from 0.541 to 0.598 under identical conditions, demonstrating that explicit reasoning guidance helps the model capture structural connectivity. This improvement was preserved even with a reduced 3{,}600-sample dataset (v2.2, v3.0), effectively halving the data requirement.

\textbf{Multimodal integration (v3.1, v3.2):} Incorporating grayscale shading as a secondary visual channel produced the second-largest performance leap (Evaluation Score 0.360$\to$0.611). To enhance visual-textual alignment, we introduced a standardized coordinate system and node-size encoding, enabling explicit mapping between image features and topological skeletons. Shadow-overlay inputs (v3.2) provided marginal additional gains.

\textbf{Two-stage training (v4.0--v4.2):} The reward-modulated objective further improved topological similarity above 0.83, with the best overall balance achieved in the v4.2 series (Topological Similarity 0.844, Evaluation Score 0.677).

\begin{table*}[htbp]
    \centering
    \caption{Comprehensive quantitative evaluation of representative model settings. Values are averaged across test sets.}
    \label{tab:summary_results}
    \resizebox{\textwidth}{!}{
    \begin{tabular}{@{}l@{\hspace{6pt}}c@{\hspace{6pt}}c@{\hspace{6pt}}c@{\hspace{6pt}}c@{\hspace{6pt}}c@{\hspace{6pt}}c@{\hspace{6pt}}c@{}}
        \toprule
        \textbf{Model Setting} &
        \makecell[c]{\textbf{Training}\\\textbf{Stage}} &
        \makecell[c]{\textbf{Training}\\\textbf{Steps}} &
        \makecell[c]{\textbf{Evaluation}\\\textbf{Score}} &
        \makecell[c]{\textbf{Connection}\\\textbf{Acc.}} &
        \makecell[c]{\textbf{Position}\\\textbf{Acc.}} &
        \makecell[c]{\textbf{Topological}\\\textbf{Sim.}} &
        \makecell[c]{\textbf{NodeNum}\\\textbf{Acc.}} \\
        \midrule
        Baseline(v1.0) & Single & 600 & 0.356 & 0.156 & 0.321 & 0.784 & 0.989 \\
        + Higher sampling diversity(v1.1) & Single & 600 & 0.455 & 0.150 & 0.712 & 0.783 & 0.953 \\
        + Extended training time(v1.2) & Single & 900 & 0.338 & 0.105 & 0.341 & 0.563 & 0.925 \\
        + Early stop regularization(v2.0) & Single & 300 & 0.355 & 0.104 & 0.351 & 0.541 & 0.901 \\
        + Step-by-step prompt(v2.1) & Single & 600 & 0.361 & 0.113 & 0.342 & 0.598 & 0.915 \\
        Reduced dataset (3.6k)(v2.2) & Single & 600 & 0.354 & 0.112 & 0.332 & 0.582 & 0.914 \\
        + Step-by-step prompt (small data)(v3.0) & Single & 600 & 0.360 & 0.113 & 0.341 & 0.593 & 0.915 \\
        + Multimodal integration(v3.1) & Single & 600 & 0.611 & 0.237 & 0.652 & 0.822 & 0.994 \\
        + Shadow overlay(v3.2) input & Single & 600 & 0.620 & 0.244 & 0.671 & 0.825 & 0.994 \\
        + Two-stage with line drawings(v4.0) & Two-stage & 600 & 0.631 & 0.238 & 0.721 & 0.834 & 0.994 \\
        + Two-stage with shadow overlay(v4.1) & Two-stage & 600 & 0.646 & 0.258 & 0.752 & 0.841 & 0.994 \\
        \textbf{Our best model (v4.2 series)} & \textbf{Two-stage} & \textbf{600} & \textbf{0.677} & \textbf{0.267} & \textbf{0.859} & \textbf{0.844} & \textbf{1.000} \\
        \bottomrule
    \end{tabular}
    }
\end{table*}

\textbf{Zero-shot and dataset-only ablation:} Under zero-shot settings, general-purpose VLMs (GPT-4.1, Claude Sonnet 4, Gemini 2.5 Pro, Qwen2.5-VL) failed to produce topologically valid outputs. Using our dataset alone without the Sketch2MinSurf structured encoding paradigm likewise led to negligible improvement, confirming that both the dataset and the encoding framework are essential.

\section{Extended Discussion}

\subsection{Structured encoding and topological skeleton}

The Sketch2MinSurf Description enables structured and differentiable encoding of node coordinates, solid and void edges, and node sizes, forming a robust parametric representation of architectural topologies. Unlike conventional boundary-based modeling~\cite{piegl2012nurbs}, this formulation operates on explicit connectivity to characterize both solid and void relations within minimal surface systems. This structure-centered data representation bridges visual sketch perception with topological reasoning, improving scalability and adaptability to high-dimensional geometric configurations.

\subsection{Reward-modulated structural loss and topology-aware surface decoding}

The S2MS-Loss modulates the supervised training objective through reward-weighted scaling, reinforcing topologically faithful skeleton generation while suppressing structural deviations. Because discrete graph operations (e.g., node matching, edge evaluation) are non-differentiable, the structural reward acts as a scalar multiplier on the loss rather than being directly differentiated through. It is important to note that the overall pipeline is not fully differentiable: the VLM produces topological skeletons through supervised fine-tuning with reward-shaped scaling, while the subsequent surface realization is performed by a non-differentiable geometric solver (Grasshopper). This decoupled design is deliberate---the solver enforces strict adherence to fabrication constraints (watertightness, curvature continuity, parametric editability) that purely neural implicit representations struggle to guarantee. Paired with the encoder, this modulated objective enables direct translation from topological skeletons to smooth minimal surfaces while preserving compatibility with Grasshopper's parametric environment. In contrast to voxel or NeRF-based generative pipelines~\cite{kim2020citycraft,tono2024vitruvio,asmar2020machinic}, the framework maintains full geometric editability and structural coherence, a requirement critical for architectural computation and fabrication.

\subsection{Spatial cognition and topological awareness in sketches}

At the spatial reasoning level, Sketch2MinSurf integrates cognitive understanding of topological connectivity rather than relying solely on pixel- or point-cloud-based representations~\cite{luo2021diffusion3dpoint,yang2019pointflow,lin2018efficientpointcloud}. The system formalizes architectural sketch logic---expressive continuity of solids and voids---into interpretable skeleton operations. Even for line drawings without shadow cues, our best model consistently achieves a topological similarity above 0.84, reflecting a strong capacity for generalization and accurate reconstruction of complex spatial configurations.

\section{Extended Conclusion}

To further enhance learning and generalization, we constructed a multimodal dataset for minimal surface generation that combines line drawings, renderings, and structured annotations. It employs a camera-based coordinate system and guided prompts, substantially improving the model's ability to interpret sketches and generalize across diverse spatial configurations. The dataset construction process systematically explored three prompt variants (Simple, Exact, Detailed) and two coordinate systems (relative, camera-based), enabling controlled ablation of each design choice.

Departing from conventional GAN, NeRF, or diffusion-based frameworks, Sketch2MinSurf encodes the differential-geometric properties of minimal surfaces as a structured language. This enables vision-language models to emulate architects' spatial reasoning---explicitly modeling continuity, hierarchy, and the interplay of solid and void. Our results demonstrate how formalizing architectural topology within a learning framework can expand AI's spatial reasoning capability and open new directions for human--AI collaboration in computational design. Overall, this work bridges design theory and computer vision, positioning topological encoding as a foundation for interpretable and generative 3D spatial modeling.

\section{Application in Architectural Practice}

The Sketch2MinSurf pipeline was deployed in a university design project to construct a full-scale minimal surface installation. Conceptual sketches were processed through the model to produce editable 3D topological skeletons, refined under site-specific constraints using dynamic relaxation in Grasshopper, and physically fabricated as a spatial gridshell structure with integrated lighting. This demonstrates the system's capacity to translate conceptual sketches into real constructed artifacts satisfying both design intent and structural logic.

\begin{figure}[htbp]
  \centering
   \includegraphics[width=1.0\linewidth]{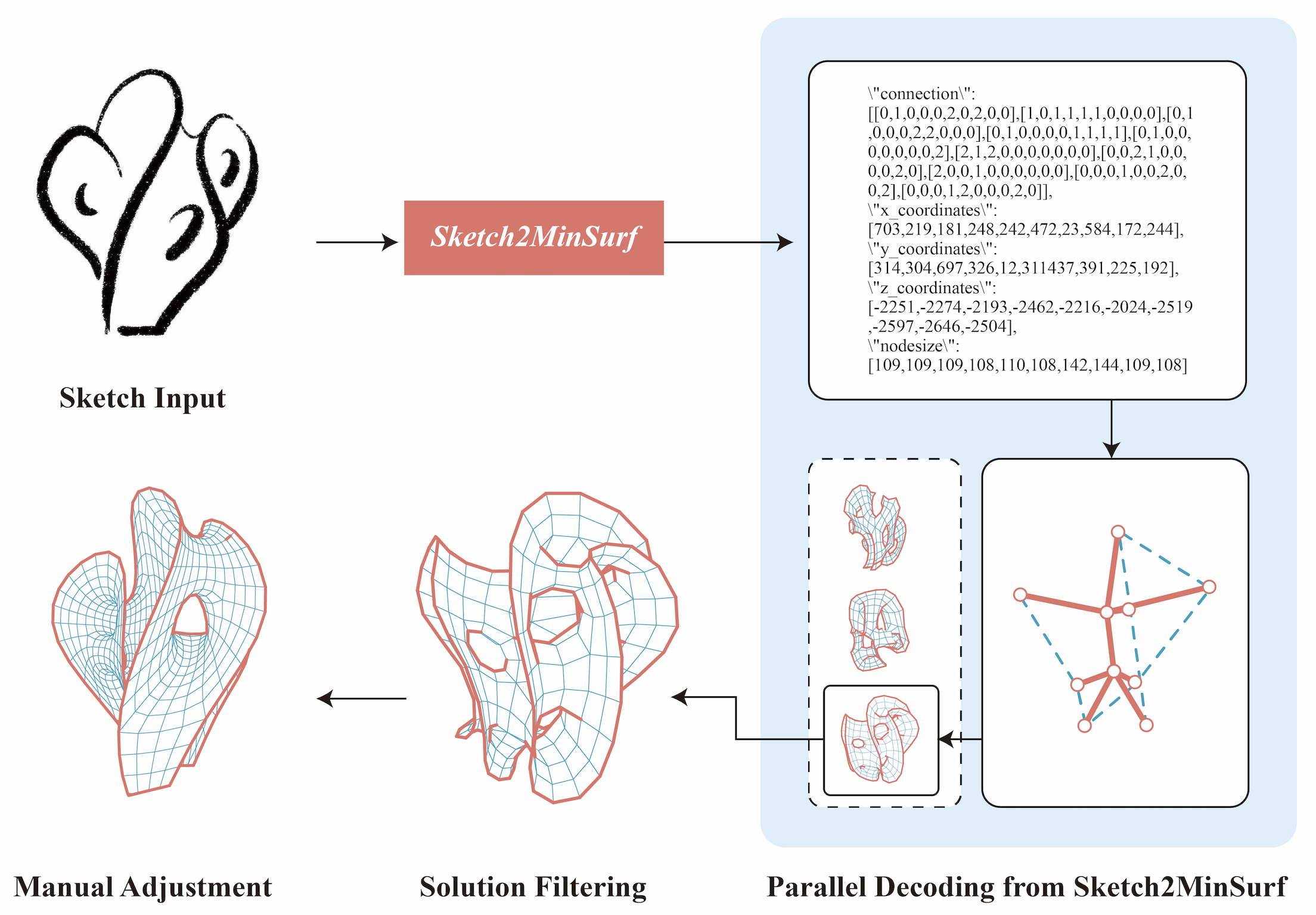}
   \caption{Complete end-to-end design-to-construction workflow using Sketch2MinSurf.}
   \label{fig:process_supp}
\end{figure}

\begin{figure}[htbp]
  \centering
  \begin{subfigure}{0.45\linewidth}
    \centering
    \includegraphics[width=1\linewidth]{figure/fig_16_1.jpeg}
    \caption{Daytime view of built installation (on-site).}
    \label{fig:daytime_supp}
  \end{subfigure}
  \hfill
  \begin{subfigure}{0.45\linewidth}
    \centering
    \includegraphics[width=1\linewidth]{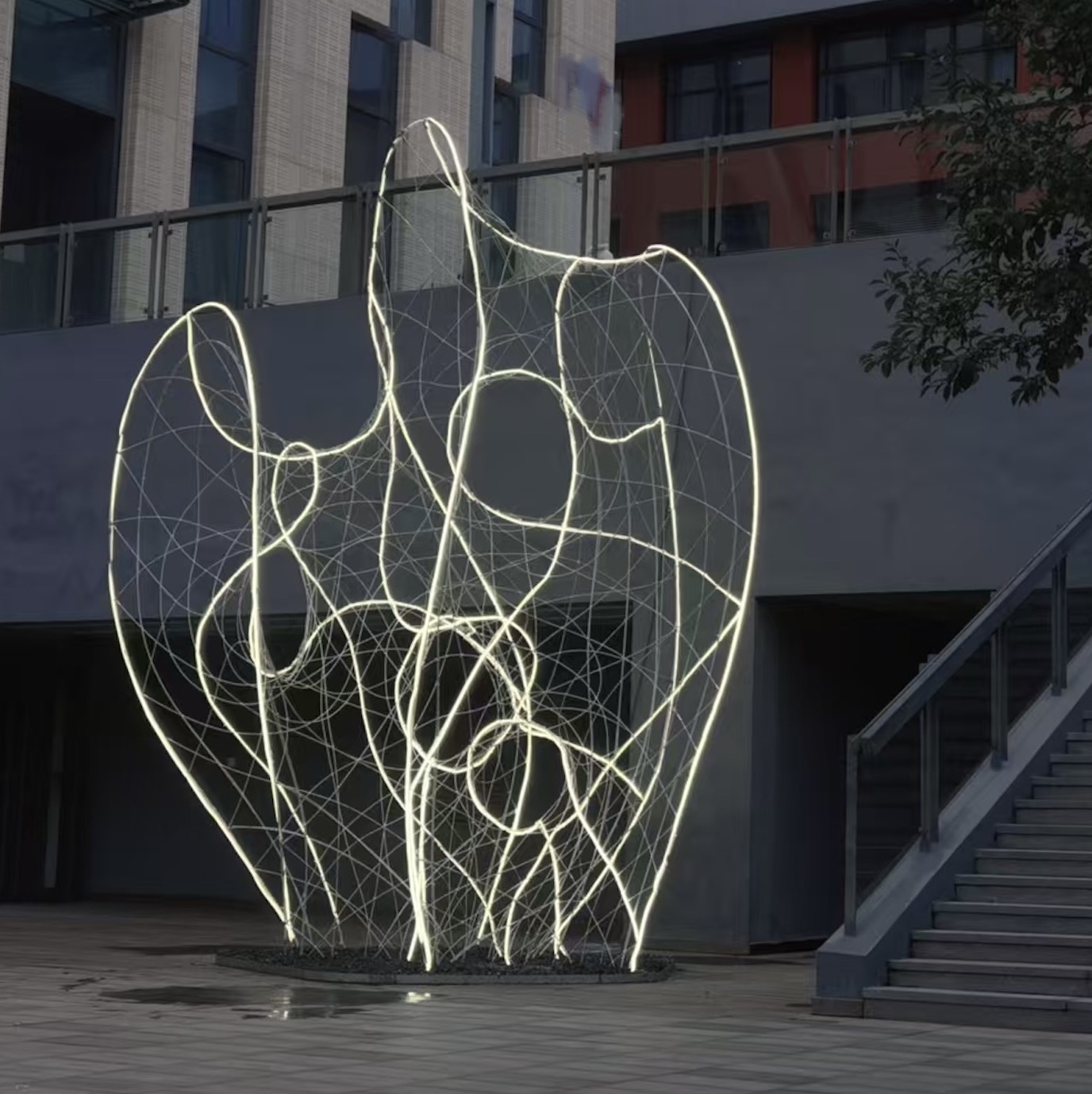}
    \caption{Nighttime illumination effect.}
    \label{fig:nighttime_supp}
  \end{subfigure}
  \caption{Additional photographs of the completed Sketch2MinSurf-based architectural installation (constructed and documented on-site, August 2025).}
  \label{fig:photograph_supp}
\end{figure}

Fig.~\ref{fig:process_supp} illustrates the complete end-to-end workflow: from initial hand-drawn sketch through VLM-based skeleton prediction, Grasshopper-based surface reconstruction and structural optimization, to final physical fabrication. The built installation (Fig.~\ref{fig:photograph_supp}) validates that the generated minimal surface geometry is not only topologically correct but also structurally feasible and aesthetically faithful to the original design intent.

\end{document}